%% file: acl_latex.tex
\pdfoutput=1

\documentclass[11pt]{article}

\usepackage[final]{acl}
\usepackage{mathtools}
\newcommand{\norm}[1]{\left\lVert #1 \right\rVert_2}
\newcommand{\cosim}{\operatorname{cos}}

\usepackage{times}
\usepackage{subcaption}
\usepackage{latexsym}
\usepackage{booktabs}
\usepackage{tabularx}
\usepackage{enumitem}

\usepackage[T1]{fontenc}

\usepackage[utf8]{inputenc}
\usepackage{amsmath}
\usepackage{amssymb}
\usepackage{amsfonts}
\usepackage{bm}
\usepackage{algorithm}
\usepackage{algorithmicx}
\usepackage{algpseudocode}
\usepackage{wrapfig}
\usepackage{bbding}

\usepackage{multirow}
\usepackage{xcolor}
\usepackage{graphicx}
\usepackage{caption}

\usepackage{microtype}

\usepackage{inconsolata}

\usepackage{colortbl}
\usepackage{booktabs,multirow}
\usepackage{makecell}
\usepackage{tabulary}
\usepackage{pifont}
\usepackage{tikz}
\usepackage[most]{tcolorbox}
\usepackage{courier}
\definecolor{blue1}{rgb}{0.15,0.15,0.15}
\definecolor{blue2}{rgb}{0.30,0.30,0.30}
\definecolor{blue3}{rgb}{0.45,0.45,0.45}
\definecolor{blue4}{rgb}{0.60,0.60,0.60}
\definecolor{blue5}{rgb}{0.70,0.70,0.70}
\definecolor{blue6}{rgb}{0.80,0.80,0.80}
\newtcolorbox{taskbox}[2][]{
enhanced, breakable, colframe=blue3!40, colback=blue5!5, arc=1mm, outer arc=1mm, fontupper=\small, fontlower=\small, coltitle=blue1, fonttitle=\bfseries, boxsep=1mm, left=0mm, right=0mm, top=0mm, bottom=0mm, before={\noindent}, segmentation style={solid, blue3}, title=#2, #1
}

\definecolor{tkcolor}{RGB}{224,223,255}
\newtcolorbox{takeaways}[2][]{
width=\columnwidth, colback = tkcolor, colframe = tkcolor, boxsep=0pt,left=10pt,right=10pt,top=2pt,bottom=3pt, fontupper=\linespread{0.9}\selectfont, title=#2,#1}
\newtcolorbox{mybox}[2][]{
width=\columnwidth, colback = gray!8, colframe = gray!8, boxsep=0pt,left=10pt,right=10pt,top=0pt,bottom=0pt, fontupper=\linespread{0.9}\selectfont, before upper={\setlength{\parskip}{4pt}\setlength{\parindent}{0pt}}, title=#2,#1}

\newlength\savewidth

\title{Less Languages, Less Tokens: An Efficient Unified Logic Cross-lingual Chain-of-Thought Reasoning Framework}

\author{
\textbf{Chenyuan Zhang$^{1,6}$\thanks{Equal contribution.} \quad Qiguang Chen$^{2}$\footnotemark[1] \quad Xie Chen$^{5,6}$ \quad Zhuotao Tian$^{1}$ \quad Bowen Xing$^{4}$} \\ \textbf{ Meishan Zhang$^{1}$ \quad Libo Qin$^{1,2,3}$\thanks{Corresponding author.} \quad Baotian Hu$^{1}$ \quad Min Zhang$^{1}$} \\ $^{1}$ Harbin Institute of Technology, Shenzhen \\ $^{2}$ Central South University \\ $^{3}$ Text Computing and Cognitive Intelligence Ministry of  \\ Education Engineering Research Center, Guizhou University \\ $^{4}$ University of Science and Technology Beijing \\ $^{5}$ Shanghai Jiao Tong University \\ $^{6}$ Shanghai Innovation Institute \\ \texttt{qinlibo@hit.edu.cn}	\\ \texttt{cyzhang@stu.hit.edu.cn, charleschen2333@gmail.com}
}

\begin{document}
\maketitle

\begin{abstract}
Cross-lingual chain-of-thought (XCoT) with self-consistency markedly enhances multilingual reasoning, yet existing methods remain costly due to extensive sampling of full trajectories across languages. Moreover, multilingual LLM representations vary strongly by language, hindering direct feature comparisons and effective pruning. Motivated by this, we introduce UL-XCoT, the first efficient unified logic cross-lingual reasoning framework that minimizes redundancy in token usage and latency, yielding the greatest efficiency under limited sampling budgets during inference. Specifically, UL-XCoT (1) achieves less languages by selecting, per query, a small candidate language set in a language-invariant unified logic space, (2) enables less tokens by monitoring logic-space trajectory dynamics during decoding to prune low-quality reasoning paths, and (3) aggregates the remaining high-quality trajectories via voting. Experiments on PolyMath across 18 languages and MMLU-ProX-Lite across 29 languages with DeepSeek-R1-Distill-Qwen-7B demonstrate that UL-XCoT achieves competitive accuracy while sharply cutting over 50\% decoding token cost versus prior sampling baselines. UL-XCoT also delivers more stable gains on low-resource languages, underscoring consistently superior robustness where standard XCoT self-consistency method fails.
\end{abstract}

\input{section/Introduction}

\input{section/method}
\input{section/experiments}
\input{section/related}
\input{section/conclusion}

\input{section/limitation}
\input{section/acknowledgement}

\input{acl_latex.bbl}
\clearpage
\input{section/appendix}

\end{document}

%% file: section/Introduction.tex
\section{Introduction}

\begin{figure}[t]
	\centering
	\includegraphics[width=0.48\textwidth]{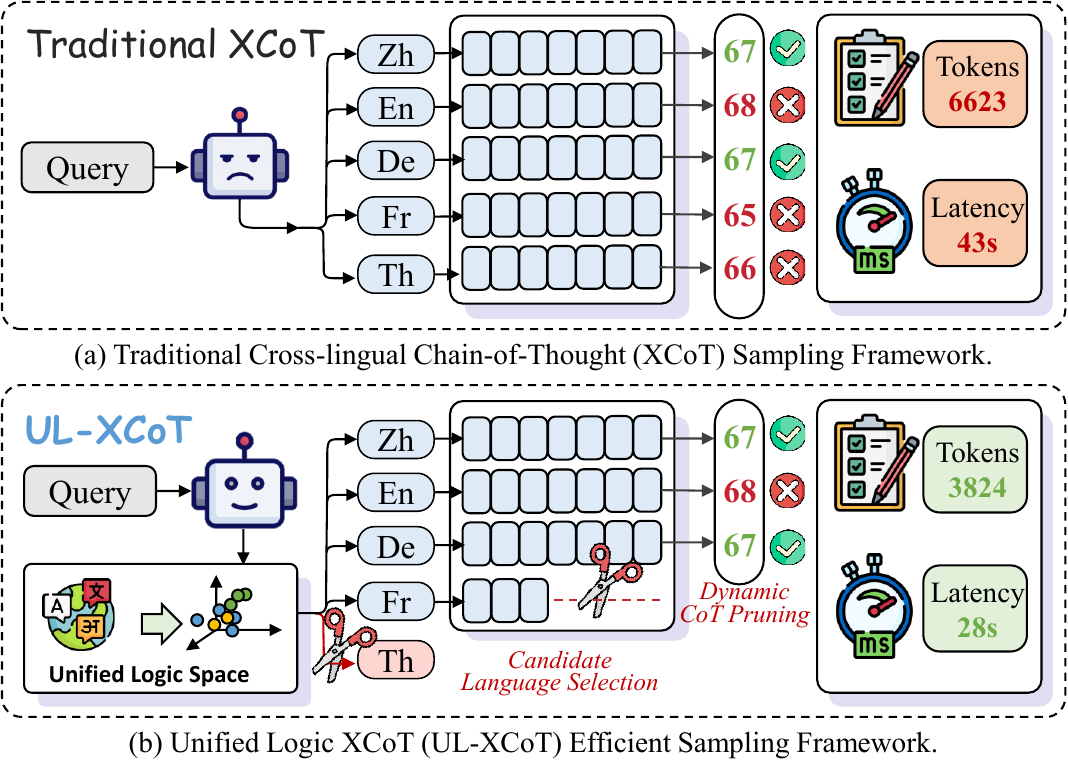}
	\caption{
		Traditional XCoT Sampling framework (a) generates complete reasoning trajectories with all languages (e.g., Chinese, English, German, French, and Thai). In contrast, Unified Logic XCoT (UL-XCoT) Efficient Sampling framework (b) uses the Unified Logic Mechanism for efficient language selection and selective trajectories generation.
	}
	\label{fig:intro}
\end{figure}

Multilingual large language models (MLLMs) have shown strong reasoning and generalization abilities~\citep{qin2025survey,chen2024breaking,lai2023chatgpt,resck2025explainability}, which Cross-lingual chain-of-thought (XCoT) can further optimize. Specifically, XCoT is a reasoning paradigm where inputs and intermediate steps use different languages~\citep{qin2023cross,zhang2024autocap,huang2024survey,tran2025disentangling}, effectively activating their core reasoning capabilities~\citep{ahuja2025efficientxlang,chen2024unlocking,huang2023not}.

As multilingual reasoning research advances, XCoT has attracted growing interest for test-time scaling via self-consistency, which samples multiple complete trajectories and aggregates outputs~\citep{tran2025scaling,khairi2025life,ghosh2025multilingual,zhang2025survey}. Specifically, \citet{qin2023cross} use voting across multiple XCoT trajectories to exploit cross-lingual complementarities. \citet{zhang2024autocap} optimize instructions to select matching languages and assign language-specific voting weights during aggregation. \citet{ranaldi2024tree} prompt the model to simulate multilingual experts, generate diverse trajectories, and derive answers via cross-referencing. \citet{khairi2025life} apply temperature scaling and Bayesian risk to sample and select high-quality XCoT trajectories.

While existing test-time scaling strategies on XCoT have improved reasoning consistency and performance, they remain limited by computational inefficiency. As shown in Figure~\ref{fig:intro}(a), most prior approaches suffer from two drawbacks: (1) \textbf{full-language sampling}, which requires generating all candidate languages, and (2) \textbf{full-trace reasoning}, which requires generating all complete reasoning paths during inference. These issues cause redundant computation on ineffective languages or similar reasoning trajectories, leading inference costs to grow linearly with the number of languages and producing substantial redundant tokens.

Motivated by this, we propose the \textbf{Unified Logic Cross-lingual Chain-of-Thought (UL-XCoT) self-consistency framework}, which enhances XCoT efficiency via two modules: Candidate Language Selection (CLS) and Dynamic CoT Pruning (DCP). As shown in Figure \ref{fig:intro} (b),
{UL-XCoT first utilizes a Unified Logic Mechanism (ULM) to establish a unified logical representation space to compare and filter reasoning processes across languages on a shared scale. It then applies CLS to evaluate candidate languages in this space and select a small subset \textbf{(less languages)} most relevant to the input query, reducing computation from irrelevant ones. Next, during reasoning, DCP tracks each language's CoT evolution and dynamically prunes redundant paths \textbf{(less tokens)} that are logically inconsistent. Finally, voting on the remaining high-quality cross-lingual reasoning paths cuts costs while preserving reasoning quality.}

Experiments on PolyMath across 18 languages and on MMLU-ProX-Lite across 29 languages show that UL-XCoT performs an obvious accuracy-efficiency trade-off with DeepSeek-R1-Distill-Qwen-7B. On PolyMath, it attains competitive difficulty-weighted accuracy while consistently requiring the fewest generated tokens and the lowest latency across languages, reducing the average token count by more than 50\% relative to AUTOCAP and by more than 65\% relative to SC. Beyond mathematical reasoning, UL-XCoT also generalizes well to MMLU-ProX-Lite, where it improves average accuracy while retaining clear efficiency advantages in both token usage and latency. Moreover, UL-XCoT yields stronger and more stable gains on a data-driven low-resource language subset, indicating greater robustness when standard prompting and sampling signals are weak.

Overall, the contributions of the paper are summarized as follows:
\begin{itemize}[leftmargin=16pt,topsep=0pt,itemsep=0pt]

\item We first point out an inherent efficiency limitation in the previous cross-lingual ensemble reasoning paradigm, where full-language enumeration implicitly assumes that the reasoning process in each language is equally important and must be fully generated, leading to substantial redundant computation.

\item We introduce UL-XCoT, an efficient XCoT self-consistency framework improving inference efficiency from two dimensions: (1) less languages through efficient language selection, and (2) less tokens through dynamic pruning with early stopping during reasoning.

\item Experimental results show that UL-XCoT significantly reduces inference cost while preserving accuracy, and yields particularly strong gains in low-resource languages.
\end{itemize}

For reproducibility, the code for this paper is available at \url{https://github.com/chenyuanTKCY/UL-XCoT}.

%% file: section/method.tex
\section{Method}

\begin{figure*}[!t]
    \centering
    \includegraphics[width=\textwidth]{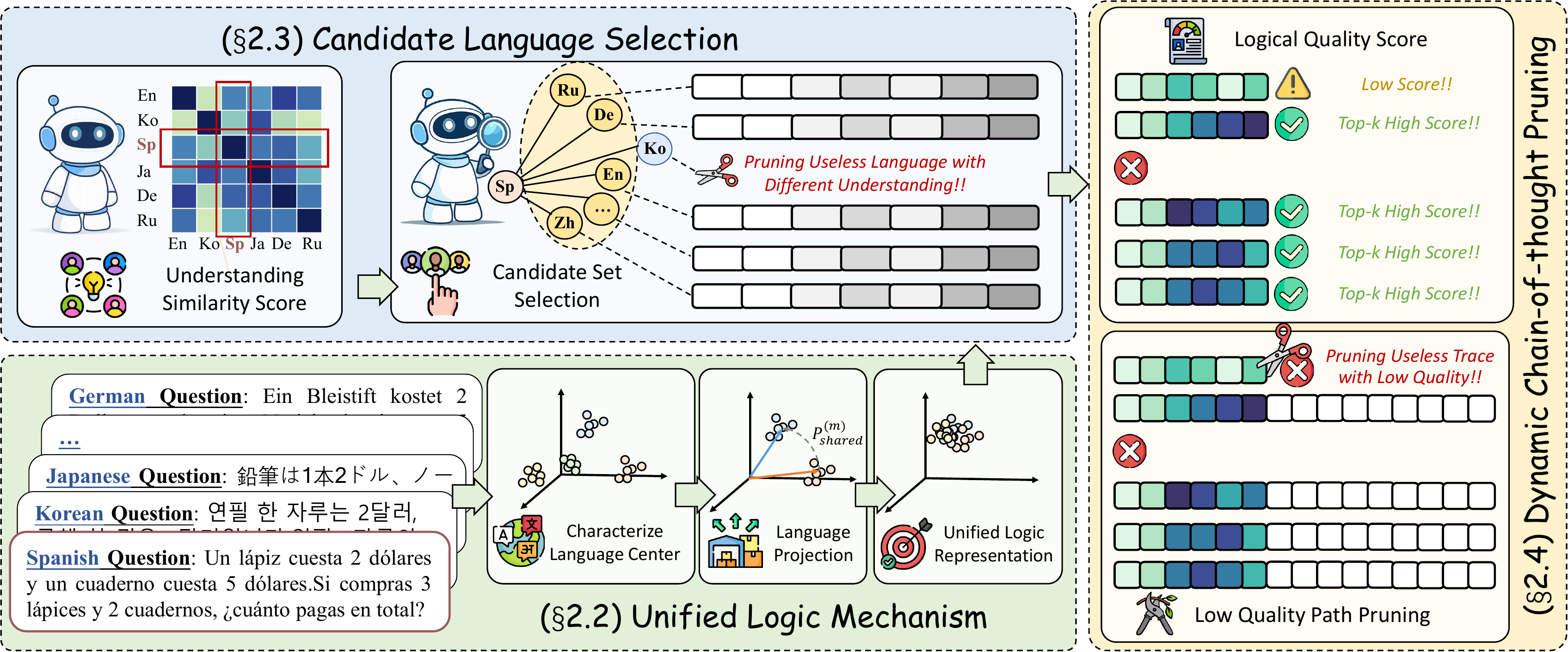}
    \vspace{-7mm}
    \caption{{Overall framework of UL-XCoT}, containing (i) The Unified Logic Mechanism, (ii) Candidate Language Selection, (iii) Dynamic Chain-of-Thought Pruning modules.}
    \vspace{-5mm}
\label{fig:overview}
\end{figure*}

In this section, we illustrate UL-XCoT. As illustrated in Figure~\ref{fig:overview}, given an input query $x$ written in language $\ell$, UL-XCoT introduces a \textbf{Unified Logic Mechanism} that makes reasoning states across different languages {comparable and measurable}.

\subsection{Overall Pipeline}
Formally, let $\mathcal{L}=\{\ell_1,\ldots,\ell_M\}$ denote the set of all languages, $f_\theta$ an MLLM, and $\mathcal{A}$ the answer space. The overall inference process yields $\hat{a}=\textsc{UL-XCoT}_\theta(x|f_\theta)\in\mathcal{A}$. Specifically, this process comprises four stages:

\begin{itemize}[leftmargin=16pt,topsep=0pt,itemsep=0pt]

\item First, in order to achieve cross-lingual comparability of reasoning states in a unified logic space, we construct a unified logic mechanism via a projection operator $P^{(m)}_{\text{shared}}$ at layer $m$.
\item {Second, we select a candidate languages set
$
\mathcal{L}_{\text{par}}(x),
$
guided by an understanding similarity score under the unified logic mechanism, thereby achieving less languages.}

\item {Third, for each $\ell\in\mathcal{L}_{\text{par}}(x)$, we perform XCoT decoding with $f_\theta$, generating trajectory $x_{\ell}$ token by token in parallel. During decoding, a time-evolving confidence signal dynamically prunes low-quality paths, yielding a retained subset of languages $\mathcal{S}(x)\subseteq \mathcal{L}_{\text{par}}(x)$. The remaining trajectories are denoted by $S^*(x)=\{x_{\ell}:\ell\in \mathcal{S}(x)\}$. In this way, we can achieve less tokens during decoding.}

\item Finally, we extract an answer $a_{\ell}$ from each surviving trajectory and aggregate them via a voting operator $V$:
\begin{equation}
\hat{a}=V\!\left(\{a_{\ell}:\ell\in \mathcal{S}(x)\}\right).
\end{equation}

\end{itemize}
Below, we describe: (1) the Unified Logic Mechanism (ULM) for constructing the unified logic space, (2) Candidate Language Selection (CLS) for forming $\mathcal{L}_{\text{par}}(x)$, and (3) Dynamic Chain-of-Thought Pruning (DCP) for obtaining $S^*(x)$.

\subsection{The Unified Logic Mechanism}
\label{sec:confidence}

To fairly compare reasoning behaviors across languages, we need a unified representation space that suppresses language-specific surface variations while preserving task-relevant reasoning structures. We thus construct a \textbf{unified logic space}, where the signals act as decision criteria for CLS (Sec.~\ref{sec:prefilter}) and DCP (Sec.~\ref{sec:pruning}).

Motivated by evidence that latent representations encode both linguistic structure and reasoning-relevant signals~\citep{zhao2025less,hao2025traininglargelanguagemodels, feng2024monitoringlatentworldstates},  we use the transformation of embeddings for constructing the Unified Logic Mechanism. Let $H_{m}(x_\ell)\in\mathbb{R}^d$ denote the hidden state at Transformer layer $m$ for sample $x_\ell$ expressed under language $\ell\in\mathcal{L}$. To characterize systematic language-dependent shifts, we use a fixed {validation set} $\mathcal{X}_{\text{val}}=\{x^i_{\ell_i}\}_{i=1}^{N}$ and obtain language-specific realizations of the {same} content for each $\ell$ (e.g., by translation or consistent prompting). We then define a {language center} at layer $m$ as
\begin{equation}
\mu^{(m)}_\ell=\frac{1}{|\mathcal{X}_{\text{val}}|}\sum_{x^i_\ell\in \mathcal{X}_{\text{val}}} H_{m}(x^i_\ell).
\end{equation}
Stacking centers across languages yields a multilingual shift matrix
\begin{equation}
M^{(m)}=\big[\mu^{(m)}_{\ell_1},\,\mu^{(m)}_{\ell_2},\,\ldots,\,\mu^{(m)}_{\ell_{|\mathcal{L}|}}\big]\in\mathbb{R}^{d\times|\mathcal{L}|}.
\end{equation}
Following the intuition that cross-lingual variation concentrates on a low-dimensional subspace, we extract its principal directions via SVD:
\begin{equation}
M^{(m)}=U^{(m)}\Sigma^{(m)}V^{(m)\top}.
\end{equation}
Let $r$ be a hyperparameter. We take the top-$r$ left singular vectors as an orthonormal basis of the language-variation subspace:
\begin{equation}
B^{(m)}_{\text{lang}}=U^{(m)}_{:,1:r}.
\end{equation}
Its orthogonal complement defines the cross-lingually shared subspace, with projection operator
\begin{equation}
P^{(m)}_{\text{shared}} = I - \lambda B^{(m)}_{\text{lang}} B^{(m)\top}_{\text{lang}}.
\end{equation}
For any input, we obtain its unified-logic-space representation by projection:
\begin{equation}
\tilde{H}_{m}(x_\ell)=P^{(m)}_{\text{shared}}h^{(m)}(x_\ell).
\end{equation}

\subsection{Candidate Language Selection}
\label{sec:prefilter}

Candidate Language Selection (CLS) compares inputs across languages in the unified logic space for pre-screening. Since it requires no generation, hidden states in this space capture a unified understanding of the input. We compute an understanding similarity score for each language and select the top-$k$ as screening candidates.

\paragraph{Understanding Similarity Score}
To quantify cross-lingual understanding consistency, we define an understanding similarity score in the unified logic space. For input query $x_\ell$ in source language $\ell$ and each candidate target language $\ell'\in\mathcal{L}$, we construct a semantically equivalent rendition $x_{\ell'}$. At analysis layer $m=a$, we extract the last-token projected representations $\tilde{H}_{a}(x_\ell)$ and $\tilde{H}_{a}(x_{\ell'})$ in the unified logic space as the model's understanding states. We define the Understanding Similarity Score (USS) as:
\begin{equation}
    \mathrm{USS}(x_\ell, x_{\ell'})\triangleq
\frac{\langle \tilde{H}_{a}(x_\ell),\,\tilde{H}_{a}(x_{\ell'})\rangle}
{\|\tilde{H}_{a}(x_\ell)\|_2\ \|\tilde{H}_{a}(x_{\ell'})\|_2}.
\end{equation}
This metric quantifies the preservation of identical understanding states across languages.

\paragraph{Candidate Set Selection}
We select the top-$k$ languages according to this score:
\begin{equation}
\mathcal{L}_{\text{par}}(x_\ell)=\text{Top-}k_{\ell'\in\mathcal{L}}\, \mathrm{USS}(x_\ell, x_{\ell'}).
\end{equation}
It performs parallel XCoT sampling only over $\mathcal{L}_{\text{par}}(x_\ell)$. By enabling direct cross-lingual comparison in the unified logic space, this pre-filtering step is query-adaptive and reduces redundant tokens generated while preserving effective cross-lingual collaboration during inference.

\subsection{Dynamic Chain-of-Thought Pruning}
\label{sec:pruning}

Dynamic Chain-of-Thought Pruning (DCP) monitors reasoning during XCoT decoding within the unified logic space and prunes low-quality paths online. This eliminates redundant generation from inconsistent or drifting trajectories, focusing computation on coherent high-quality paths.

\paragraph{Logical Quality Score}
We introduce a warm-up phase of length $T_{\text{warm}}$, during which pruning is disabled to avoid discarding paths prematurely. After warm-up, let $S_t$ denote the reasoning path from 0 to $t$-th step. For each path in $S_t$ with language $\ell$, we track its trajectory and compute a cohort score ``$\texttt{score}^{(t)}(\cdot|\ell)$'' from $t_0$-th step  over a window of length $\tau$. The Logical Quality Score (LQS) is\footnote{Details are in Appendix~\ref{app:math}.}:
\begin{equation}
\mathrm{LQS}(S_t|x_\ell,\ell')\triangleq \int_{t=t_0}^{t_0+\tau} \texttt{score}(S_t|x_\ell,\ell').
\end{equation}

\paragraph{Low Quality Path Pruning.}
At the end of the monitoring window ($T_{E}=T_{\mathrm{warm}}+\tau+1$), we collect the trace set $\mathcal{S}$ by retaining the top- $k'$ paths ranked by $\mathrm{LQS}(\ell)$:
\begin{equation}
\mathcal{S}(x_\ell) \;=\; \text{Top-}k^{'}_{\ell'\in \mathcal{L}_{\text{par}}(x_\ell)}\ \mathrm{LQS}(S_t|x_\ell,\ell'),
\end{equation}
where $k'$ follows pruning ratio $\rho$. Remaining paths terminate early, while retained paths continue decoding and aggregate via voting. This adaptively allocates compute to coherent paths, reducing redundancy and ensuring robustness.

%% file: section/experiments.tex
\section{Experiments}

\begin{figure*}[ht]
  \centering
  \includegraphics[width=\textwidth]{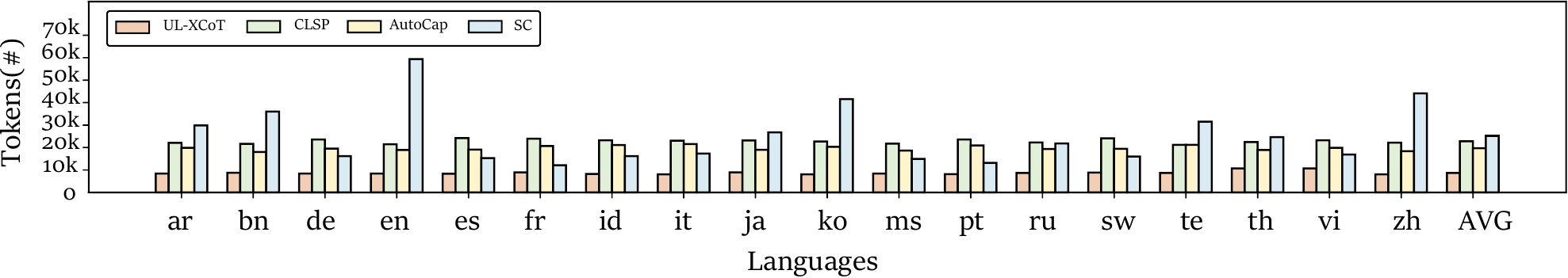}
  \vspace{-8mm}
  \caption{Average decoding token cost during generation on PolyMath.  }
  \vspace{-1mm}
\label{fig:avg_tokens}

\end{figure*}
\begin{figure*}[ht]
  \centering
  \includegraphics[width=\textwidth]{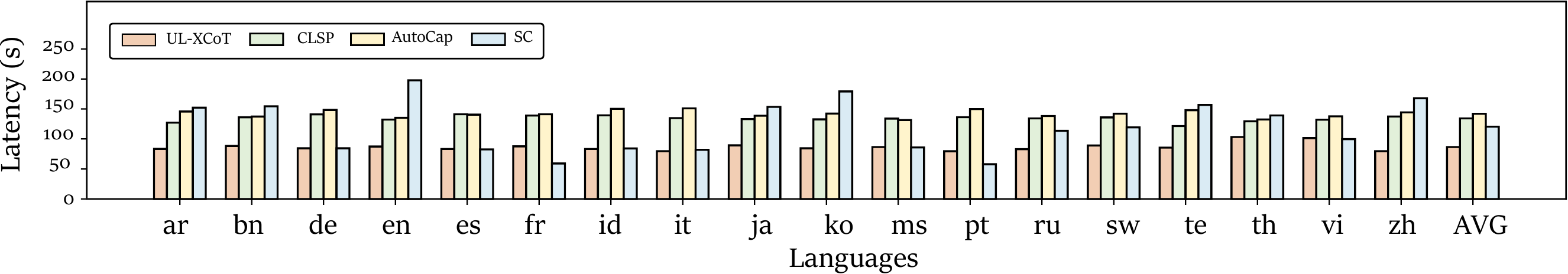}
  \vspace{-8mm}
  \caption{Average end-to-end latency across languages during generation on PolyMath.}
  \vspace{-1mm}
\label{fig:avg_latency}
\end{figure*}

\subsection{Experimental Setting}

\label{subsec:Setting}

\paragraph{Benchmark.}
We mainly evaluate on PolyMath~\cite{Wang2025}, a multilingual mathematical reasoning benchmark with {18 parallel languages} and {4 difficulty levels}.

To test generalization, we further select a complementary benchmark on MMLU-ProX-Lite~\citep{xuan-etal-2025-mmlu}, a multilingual multiple-choice benchmark spanning broader knowledge and reasoning categories.

\paragraph{Evaluation protocol.}
We assess both effectiveness and efficiency: (1) \textbf{Accuracy:} All methods use the same backbone and decoding constraints; for SC and AUTOCAP, we control the sample budget at UL-XCoT's worst-case level. We follow PolyMath in reporting DW-ACC. (2) \textbf{Efficiency:} We measure inference cost via generated tokens and wall-clock latency under identical conditions.
\paragraph{Experimental setup.}
We conduct all experiments with {DeepSeek-R1-Distill-Qwen-7B}~\cite{guo2025deepseek}. All runs are executed on NVIDIA RTX A6000 GPUs (48\,GB). We set the maximum generation length in the interval of 2048--10240 according to PolyMath difficulty and stability of its performance. For prompting, we use a \textbf{concise-reasoning template}\footnote{The prompt is provided in Appendix~\ref{app:prompt}.} for our method and baseline.

\paragraph{UL-XCoT configuration.}
We set $\lambda=0.4$, use layer $m=13$ for logic-space representations (Section~\ref{sec:prefilter}), set $|\mathcal{L}_{\text{par}}(x)|=9$--12, and use a warm-up stage with $T_{\text{warm}}=10$. After warm-up, we compute trajectory signals in a sliding window $\tau=3c$ and adjust the pruning ratio $\rho$ within 20\%--60\% throughout decoding.

\paragraph{Baselines.}
We compare UL-XCoT with representative reasoning strategies spanning single-path prompting and sampling-based test-time scaling:
{(i) CLP (CLSP\footnote{The Self-consistency version of CLP.})}~\cite{qin2023cross}, a cross-lingual prompting framework that leverages multilingual signals to improve reasoning robustness;
{(ii) CoT}~\cite{wei2022chain}, standard single-trajectory chain-of-thought prompting;
{(iii) SC}~\cite{wang2022self}, self-consistency that samples diverse CoT trajectories and aggregates the answers via majority voting;
{(iv) AUTOCAP}~\cite{zhang2024autocap}, an adaptive variant of SC that selects languages and assigns weights for aggregation;
{(v) ST-BoN}~\citep{wang2025sampling}, an efficient sampling-based method that improves test-time scaling under a fixed budget in single-language inference;
and {(vi) UL-CoT}, a monolingual counterpart of UL-XCoT featuring the identical DCP module.

\subsection{Results for UL-XCoT}

\subsubsection{The Superior Efficiency.}
\label{sec:superior_efficiency}
Figures~\ref{fig:avg_tokens} \&~\ref{fig:avg_latency} compare the average decoding cost and end-to-end latency across languages.

\noindent \textbf{UL-XCoT can achieve marked efficiency.} As illustrated in Figure~\ref{fig:avg_tokens}, UL-XCoT consistently uses the fewest tokens across all languages, achieving the lowest average token count with over 50\% reduction relative to AUTOCAP and more than 65\% relative to SC. This efficiency is mirrored in latency metrics (Figure~\ref{fig:avg_latency}), where UL-XCoT also exhibits the lowest latency in most languages\footnote{A per-difficulty breakdown of token cost and wall-clock latency is provided in Appendix~\ref{app:efficiency}.}.

\noindent \textbf{UL-XCoT can also reduce overthinking in high-resource languages.} As shown in Figure~\ref{fig:avg_tokens}, some baselines (notably \textsc{SC}) exhibit pronounced spikes in token consumption on high resource languages like English and Chinese, suggesting occasional overthinking with excessively long reasoning traces. UL-XCoT largely avoids such extreme outliers, providing more steady inference-time behavior and significant measures toward cost for multilingual deployment.

\begin{table*}[t]
  \centering
  \caption{Performance on PolyMath across 18 languages and four difficulty levels from top to bottom.}
  \label{tab:acc_polymath_the_whole}
  \resizebox{\textwidth}{!}{
  \begin{tabular}{l *{24}{c}}
  \toprule

  \textbf{ACC} & \textbf{ar} & \textbf{bn} & \textbf{de} & \textbf{en} & \textbf{es} & \textbf{fr} & \textbf{id} & \textbf{it} & \textbf{ja} & \textbf{ko} & \textbf{ms} & \textbf{pt} & \textbf{ru} & \textbf{sw} & \textbf{te} & \textbf{th} & \textbf{vi} & \textbf{zh} & \textbf{AVG} \\
  \midrule
  \rowcolor[rgb]{0.940, 0.940, 0.940}
  \multicolumn{20}{c}{PolyMath-Low}\\
  \midrule
  \texttt{CoT}~\citep{wei2022chain}
  & 52.0          & 42.4          & 56.0          & 83.2          & 72.8          & 56.8          & 59.2          & 68.0          & 41.6          & 42.4          & 55.2          & 67.2          & 71.2          & 4.0           & 13.6          & 49.6          & 59.2          & 76.8          & 54.0          \\
  \texttt{CLP}~\citep{qin2023cross}
  & 39.2          & 32.8          & 45.6          & 58.4          & 48.0          & 47.2          & 38.4          & 46.4          & 40.8          & 32.8          & 47.2          & 52.0          & 36.0          & 40.8          & 32.8          & 45.6          & 40.8          & 40.8          & 42.5           \\
  \midrule
  \texttt{SC}~\citep{wang2022self}
  & 68.8          & 59.2          & 68.8          & \textbf{88.8} & 77.6          & 76.8          & 73.6          & 78.4          & 64.8          & 64.0          & 72.0          & 73.6          & 79.2          & 12.0          & 26.4          & 66.4          & 70.4          & 81.6          & 66.8          \\

  \texttt{CLSP}~\citep{qin2023cross}
  & 77.6          & \textbf{83.2} & 81.6          & 81.6          & 80.8          & \textbf{84.8} & 82.4          & 81.6          & 82.4          & 80.8          & 78.4          & 81.6          & 80.0          & 80.0          & 80.8          & 83.2          & \textbf{84.0} & 80.0          & 81.4          \\
  \texttt{AUTOCAP}~\citep{zhang2024autocap}
  & 80.0          & 81.6          & 81.6          & 76.8          & \textbf{84.0} & 81.6          & 80.8          & 79.2          & 81.6          & 74.4          & 84.8          & \textbf{83.2} & 81.6          & \textbf{84.0} & 80.8          & 78.4          & 83.2          & 79.2          & 80.9          \\
  \texttt{ST-BoN}~\citep{wang2025sampling}
  & 60.0 & 62.4 & 65.6 & 69.6 & 64.8 & 72.0 & 64.0 & 63.2 & 61.6 & 61.6 & 63.2 & 62.4 & 64.8 & 60.0 & 63.2 & 59.2 & 61.6 & 63.2 & 63.5 \\
  \rowcolor[rgb]{0.929, 0.961, 0.980}
  \texttt{UL-CoT}
  & 68.0 & 55.2 & 68.0 & 86.4 & 79.2 & 76.0 & 71.2 & 79.2 & 63.2 & 62.4 & 73.6 & 75.2 & 73.6 & 9.6 & 24.0 & 61.6 & 72.0 & 80.0 & 65.5 \\
  \rowcolor[rgb]{0.850, 0.900, 0.940}
  \texttt{\textbf{UL-XCoT}}
  & \textbf{81.6} & \textbf{83.2} & \textbf{84.0} & 84.8          & 82.4          & 84.0          & \textbf{85.6} & \textbf{83.2} & \textbf{84.8} & \textbf{84.0} & \textbf{85.6} & \textbf{83.2} & \textbf{85.6} & 83.2          & \textbf{84.0} & \textbf{84.0} & 83.2          & \textbf{82.4} & \textbf{83.8} \\
  \midrule
  \rowcolor[rgb]{0.940, 0.940, 0.940}
  \multicolumn{20}{c}{PolyMath-Medium}\\
  \midrule
  \texttt{CoT}~\citep{wei2022chain}
  & 18.4          & 12.8          & 18.4          & 21.6          & 20.0          & 14.4          & 15.2           & 16.8         & 14.4          & 20.0          & 13.6 & 16.8         & 18.4 & 8.8 & 8.0 & 14.4 & 19.2 & 24.8 & 16.4 \\
  \texttt{CLP}~\citep{qin2023cross}
  & 17.6          & 17.6          & 16.8          & 18.4          & 18.4          & 20.0          & 15.2           & 17.6         & 20.8          & 19.2          & 16.0 & 17.6         & 18.4 & 14.4 & 14.4 & 8.8 & 21.6 & 16.8 & 17.2 \\
  \midrule
  \texttt{SC}~\citep{wang2022self}
  & 30.4          & 22.4          & 27.2          & {38.4} & 25.6          & 24.8          & 22.4           & 24.0         & 19.2          & 32.0          & 17.6 & 24.8         & 32.0        & 12.0 & 12.8 & 24.0                        & 29.6          & {39.2} & 25.5 \\
  \texttt{CLSP}~\citep{qin2023cross}
  & 32.0          & 28.8          & 33.6          & 29.6          & 28.0          & \textbf{32.8} & 31.2           & 25.6         & \textbf{31.2} & 28.8          & 30.4 & \textbf{28.8} & 28.8         & 32.0 & 30.4 & \textbf{29.6}               & \textbf{33.6} & 32.0 & 30.4   \\
  \texttt{AUTOCAP}~\citep{zhang2024autocap}
  & \textbf{35.2} & \textbf{34.4} & 30.4          & 24.0          & 31.2          & 29.6          & 25.6           & 26.4          & \textbf{31.2} & \textbf{36.0} & 29.6 & \textbf{28.8} & 27.2         & 30.4 & 28.0 & 25.6                         & 28.8        & 26.4 & 29.4 \\
  \texttt{ST-BoN}~\citep{wang2025sampling}
  & 20.8 & 15.2 & 15.2 & 19.2 & 22.4 & 16.0 & 17.6 & 17.6 & 17.6 & 16.8 & 19.2 & 20.0 & 14.4 & 16.8 & 17.6 & 24.0 & 19.2 & 21.6 & 18.4 \\
  \rowcolor[rgb]{0.929, 0.961, 0.980}
  \texttt{UL-CoT}
  & 28.0 & 25.6 & 26.4 & \textbf{45.6} & 26.4 & 24.0 & 22.4 & 25.6 & 21.6 & \textbf{36.0} & 17.6 & 19.2 & 32.0 & 13.6 & 15.2 & 27.2 & 24.0 & \textbf{43.2} & 26.3 \\
  \rowcolor[rgb]{0.850, 0.900, 0.940}
  \texttt{\textbf{UL-XCoT}}
  & 28.8          & 28.0          & \textbf{35.2} & 32.0          & \textbf{35.2} & 29.6           & \textbf{33.6} & \textbf{28.8} & 27.2           & 27.2         & 31.2 & 28.0          & \textbf{32.8} & \textbf{38.4} & \textbf{33.6} & \textbf{29.6} & 32.8 & 28.0 & \textbf{31.1} \\
  \midrule
  \rowcolor[rgb]{0.940, 0.940, 0.940}
  \multicolumn{20}{c}{PolyMath-High}\\
  \midrule
  \texttt{CoT}~\citep{wei2022chain}
  & 5.6 & 6.4 & 10.4 & 8.0 & 10.4 & 8.0 & 4.0 & 9.6 & 7.2 & 8.0 & 4.8 & 8.8 & 8.0 & 0.8 & 1.6 & 7.2 & 8.8 & 12.8 & 7.2 \\
  \texttt{CLP}~\citep{qin2023cross}
    & 7.2 & 8.0 & 8.8 & 7.2 & 6.4 & 10.4 & 8.8 & 7.2 & 12.0 & 4.8 & 9.6 & 10.4 & 8.0 & 8.0 & 6.4 & 6.4 & 6.4 & 6.4 & 7.9 \\
  \midrule
  \texttt{SC}~\citep{wang2022self}
    & 14.4 & 11.2 & 9.6 & \textbf{20.0} & 12.0 & 12.0 & 8.0 & 13.6 & 8.8 & \textbf{15.2} & 8.0 & 9.6 & 10.4 & 2.4 & 4.0 & 11.2 & 9.6 & {20.8} & 11.2 \\
  \texttt{CLSP}~\citep{qin2023cross}
    & \textbf{15.2} & {13.6} & \textbf{15.2} & 15.2 & \textbf{16.0} & 15.2 & 16.0 & \textbf{15.2} & \textbf{16.8} & 13.6 & \textbf{15.2} & \textbf{16.8} & \textbf{16.0} & \textbf{16.8} & 12.8 & 11.2 & 14.4 & 11.2 & \textbf{14.8} \\
  \texttt{AUTOCAP}~\citep{zhang2024autocap}
    & 13.6 & {13.6} & 13.6 & 14.4 & 12.8 & {20.0} & 16.8 & 14.4 & \textbf{16.8} & 12.8 & 12.0 & 14.4 & 14.4 & 12.0 & \textbf{13.6} & 12.8 & \textbf{15.2} & 12.8 & 14.2 \\
  \texttt{ST-BoN}~\citep{wang2025sampling}
  & 12.0 & 10.4 & 8.8  & 3.2  & 7.2  & 6.4  & 8.0  & 14.4 & 7.2  & 8.8  & 10.4 & 8.0  & 9.6  & 10.4 & 7.2  & 8.8  & 8.0  & 8.0  & 8.7  \\
  \rowcolor[rgb]{0.929, 0.961, 0.980}
  \texttt{UL-CoT}
  & 12.0 & \textbf{16.0} & 10.4 & \textbf{29.6} & 14.4 & 7.2 & 7.2 & 12.0 & 8.8 & 17.6 & 8.0 & 10.4 & 13.6 & 4.8 & 4.0 & 11.2 & 8.8 & \textbf{25.6} & 12.3 \\
  \rowcolor[rgb]{0.850, 0.900, 0.940}
  \texttt{\textbf{UL-XCoT}}
    & 13.6 & 11.2 & 12.8 & 11.2 & 14.4 & 15.2 & \textbf{17.6} & 13.6 & 12.8 & 11.2 & 12.8 & 14.4 & 15.2 & 14.4 & 12.0 & \textbf{16.8} & 12.8 & 12.0 & 13.6 \\
  \midrule
  \rowcolor[rgb]{0.940, 0.940, 0.940}
  \multicolumn{20}{c}{PolyMath-Top}\\
  \midrule
  \texttt{CoT}~\citep{wei2022chain}
  & 5.6 & 1.6 & 7.2 & 2.4 & 7.2 & 9.6 & 7.2 & 3.2 & 8.8 & 3.2 & 4.8 & 8.0 & 5.6 & 0.8 & 0.8 & 6.4 & 4.8 & 4.0 & 5.1 \\
  \texttt{CLP}~\citep{qin2023cross}
  & 4.0 & 2.4 & 8.0 & 9.6 & 5.6 & 8.8 & 5.6 & 8.0 & 9.6 & 8.8 & 8.8 & 4.8 & 9.6 & 8.0 & 6.4 & 4.0 & 7.2 & 8.8 & 7.1 \\
  \midrule
  \texttt{SC}~\citep{wang2022self}
  & \textbf{12.0} &  4.8 &  7.2 &  7.2 &  8.8 &  9.6 &  8.8 & \textbf{10.4} &  4.0 & 12.8 &  9.6 &  8.8 &  9.6 &  3.2 &  3.2 &  7.2 &  8.0 & {12.0} &  8.2 \\
  \texttt{CLSP}~\citep{qin2023cross}
  &  8.8 &  7.2 & \textbf{12.8} & 11.2 &  6.4 &  8.8 & \textbf{12.8} &  7.2 & \textbf{10.4} & \textbf{14.4} &  9.6 &  5.6 &  8.0 &  9.6 & \textbf{11.2} & \textbf{11.2} &  8.8 &  9.6 &  9.6 \\
  \texttt{AUTOCAP}~\citep{zhang2024autocap}
  &  8.8 & 8.8 &  8.8 &  9.6 & {9.6} &  8.8 & 10.4 &  6.4 &  7.2 & 12.8 & \textbf{14.4} & \textbf{9.6} &  6.4 &  8.0 &  9.6 &  5.6 &  8.0 & 11.2 &  9.1 \\
  \texttt{ST-BoN}~\citep{wang2025sampling}
  & 7.2  & 6.4  & 5.6  & 8.8  & 5.6  & \textbf{10.4} & 8.8  & 7.2  & 10.4 & 8.8  & 9.6  & 9.6  & 7.2  & 5.6  & 7.2  & 10.4 & 10.4 & 7.2  & 8.1  \\
  \rowcolor[rgb]{0.929, 0.961, 0.980}
  \texttt{UL-CoT}
  & 10.4 & 9.6 & 5.6 & \textbf{22.4} & \textbf{10.4} & \textbf{10.4} & 7.2 & 9.6 & 7.2 & \textbf{15.2} & 5.6 & 8.8 & 9.6 & 3.2 & 5.6 & 6.4 & 8.0 & \textbf{21.1} & 9.4 \\
  \rowcolor[rgb]{0.850, 0.900, 0.940}
  \texttt{\textbf{UL-XCoT}}
  & 10.4 & \textbf{12.8} & 10.4 & 12.8 & 9.6 & \textbf{10.4} &  9.6 &  9.6 &  9.6 &  9.6 & 11.2 &  8.0 & \textbf{11.2} & \textbf{13.6} &  8.8 & \textbf{11.2} & \textbf{12.0} & 10.4 & \textbf{10.6} \\
  \bottomrule
  \end{tabular}}
\end{table*}

\begin{table}[!t]
  \centering
  \caption{Ablation study for 3 proposed modules on PolyMath-Low subset.}
  \label{tab:ablation_study}
  \footnotesize
  \setlength{\tabcolsep}{3pt}
  \renewcommand{\arraystretch}{1.00}

  \resizebox{\linewidth}{!}{
  \begin{tabular}{l c c c}
    \toprule
    \textbf{Subset} & \textbf{ACC.} & \textbf{Latency (s)} & \textbf{Tokens (\#)} \\
    \midrule
    UL-XCoT w/o CLS & 84.4 & 36.2 & 5560 \\
    UL-XCoT w/o DCP & 81.4 & 30.7 & 3893 \\
    UL-XCoT w/o ULM & 79.8 & 25.4 & 3098 \\
    UL-XCoT w/o all modules & \textbf{85.2} & 35.9 & 7518 \\

    \cellcolor[rgb]{0.929, 0.961, 0.980}UL-XCoT &
    \cellcolor[rgb]{0.929, 0.961, 0.980}83.8 &
    \cellcolor[rgb]{0.929, 0.961, 0.980}\textbf{24.6} &
    \cellcolor[rgb]{0.929, 0.961, 0.980}\textbf{3092} \\
    \bottomrule
  \end{tabular}
  }
\end{table}

\subsubsection{The Comparable Performance.}

Table~\ref{tab:acc_polymath_the_whole} \& ~\ref{tab:dw-acc_polymath} provide a comparative evaluation of UL-XCoT against a range of multilingual reasoning baselines on PolyMath\footnote{The overall performance is provided in Appendix~\ref{app:polymath}}.

\noindent \textbf{UL-XCoT achieves competitive accuracy.} Under the same evaluation setting, UL-XCoT delivers consistently strong performance across languages and four difficulty levels, ranking at or near the top on average. Table~\ref{tab:dw-acc_polymath} shows competitive DW-ACC on PolyMath-Full across all difficulties, confirming effectiveness on challenging problems. Overall, UL-XCoT provides robust gains over prior prompting and sampling-based baselines for cross-lingual mathematical reasoning.

\noindent \textbf{UL-XCoT also shows robustness in low-resource languages.} {We evaluate robustness on a data-driven low-resource subset~\citep{bandarkar2024belebele}, }performed as languages where both \textsc{CoT} and \textsc{SC} fall below their mean DW-ACC across all languages. On this challenging set, UL-XCoT remains consistently strong performance, indicating more stable and reliable cross-lingual reasoning when language resources are limited.

\noindent \textbf{The multilingual gain does not collapse to pruning alone.} To separate the effect of multilingual collaboration from the gain brought by pruning, we further introduce a monolingual control variant, \textbf{UL-CoT}. This variant keeps the same unified logic representation and dynamic pruning signals as UL-XCoT, but samples and aggregates reasoning traces only within the query language. As shown in Table~\ref{tab:acc_polymath_the_whole}, UL-XCoT consistently outperforms UL-CoT across all four PolyMath difficulty levels,

indicating that the benefit of UL-XCoT does not come only from pruning more efficiently; multilingual interaction itself contributes substantial performance gains, especially on lower-resource languages.

\begin{table*}[t]
  \centering
  \caption{Additional benchmark results on MMLU-ProX-Lite across 29 languages.}
  \label{tab:global_mmlu_lite_acc}
  \footnotesize
  \setlength{\tabcolsep}{3.6pt}
  \renewcommand{\arraystretch}{1.02}
  \resizebox{\textwidth}{!}{
  \begin{tabular}{lcccccccccccccccccccccccccccccc}
  \toprule
  \textbf{Languages} & \textbf{af} & \textbf{ar} & \textbf{bn} & \textbf{cs} & \textbf{de} & \textbf{en} & \textbf{es} & \textbf{fr} & \textbf{hi} & \textbf{hu} & \textbf{id} & \textbf{it} & \textbf{ja} & \textbf{ko} & \textbf{mr} & \textbf{ne} & \textbf{pt} & \textbf{ru} & \textbf{sr} & \textbf{sw} & \textbf{te} & \textbf{th} & \textbf{uk} & \textbf{ur} & \textbf{vi} & \textbf{wo} & \textbf{yo} & \textbf{zh} & \textbf{zu} & \textbf{AVG} \\
  \midrule
  \texttt{CLSP}~\citep{qin2023cross} & 37.3 & \textbf{44.1} & 32.2          & \textbf{44.1} & \textbf{42.4} & 39.0          & 40.7          & \textbf{42.4} & 35.6          & 37.3          & \textbf{47.5} & 37.3          & 39.0          & \textbf{44.1} & \textbf{42.4} & 35.6          & \textbf{42.4} & 44.1          & 40.7          & 39.0          & 37.3          & 42.4          & 37.3          & 37.3          & \textbf{45.8} & 42.4          & 44.1          & \textbf{42.4} & 39.0          & 40.5 \\
  \rowcolor[rgb]{0.929, 0.961, 0.980}
  \texttt{\textbf{UL-XCoT}} & \textbf{40.7} & 39.0          & \textbf{52.5} & 40.7          & 40.7          & \textbf{49.2} & \textbf{52.5} & 40.7          & \textbf{39.0} & \textbf{42.4} & 44.1          & \textbf{45.8} & \textbf{45.8} & 39.0          & 40.7          & \textbf{42.4} & \textbf{42.4} & \textbf{45.8} & \textbf{45.8} & \textbf{42.4} & \textbf{40.7} & \textbf{45.8} & \textbf{49.2} & \textbf{39.0} & 42.4          & \textbf{45.8} & \textbf{45.8} & \textbf{42.4} & \textbf{42.4} & \textbf{43.6} \\
  \bottomrule
  \end{tabular}}
\end{table*}

\begin{figure*}[t]
  \centering
  \includegraphics[width=\textwidth]{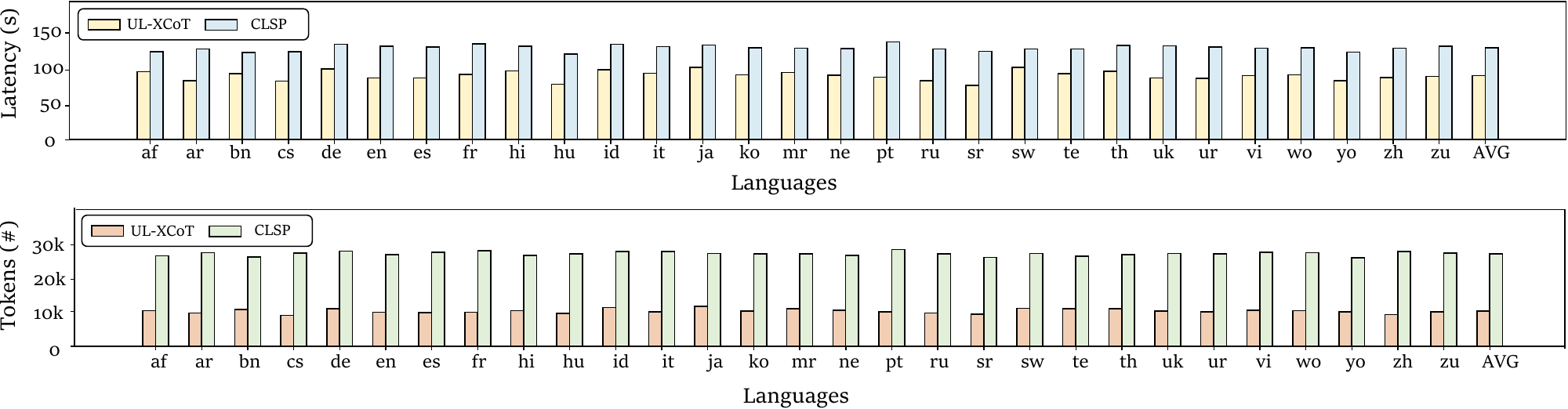}
  \vspace{-8mm}
  \caption{Average token cost and end-to-end latency across languages during generation on MMLU-ProX-Lite.  }
  \vspace{-1mm}
\label{fig:cost_mmlu}
\end{figure*}

\subsubsection{The Robust Generalization.}
\label{sec:global_mmlu_lite}

\noindent \textbf{UL-XCoT transfers beyond PolyMath.} To complement the main PolyMath evaluation, we compare UL-XCoT and CLSP on MMLU-ProX-Lite~\citep{xuan-etal-2025-mmlu} under the same evaluation protocol. This benchmark covers a broader set of multilingual knowledge-and-reasoning multiple-choice questions, providing a task setting distinct from multilingual math reasoning. As shown in Table~\ref{tab:global_mmlu_lite_acc}, UL-XCoT improves the average accuracy from 40.5 to 43.6, outperforms CLSP in 19 of 29 languages, and ties in 2 languages. And the gains also extend to several lower-resource languages, suggesting that the method can generalize across multilingual reasoning tasks.

\noindent \textbf{UL-XCoT retains its efficiency advantage on the new benchmark.} Figure~\ref{fig:cost_mmlu} reports the generated tokens and end-to-end latency across the 29 languages. UL-XCoT reduces average token usage from 27{,}679.3 to 10{,}543.6 and average latency from 134.2\,s to 93.7\,s. This result is consistent with our main findings: candidate language selection and dynamic pruning avoid unnecessary multilingual sampling, yielding a better cost--quality trade-off than CLSP even on a different task.

\subsection{Analysis}

\subsubsection{Effectiveness of Each Module}
To quantify the contribution of each module, we carry out ablation studies on PolyMath-Low

by removing one module at a time \footnote{Variants without CLS or DCP may implicitly use more compute, which can inflate accuracy; hence accuracy should be read together with compute.} in Table~\ref{tab:ablation_study}.

\paragraph{ULM is the main accuracy contributor under a matched compute budget.}
A key observation from Table~\ref{tab:ablation_study} is that {UL-XCoT and UL-XCoT w/o ULM operate under nearly identical compute budgets} (24.6s/3092 tokens vs.\ 25.4s/3098 tokens). Therefore, the accuracy drop after removing ULM reflects a {genuine algorithmic gain} rather than an artifact of increased sampling or longer decoding. Mechanistically, {ULM} maps language-specific chains into a unified, comparable logic space, enabling reliable cross-lingual confidence accumulation and coherent trajectory selection without increasing tokens and latency.

\paragraph{CLS primarily improves efficiency by reducing the search space.}
{CLS} mainly improves {efficiency} by proactively shrinking the candidate language set to a few logically
relevant participants, reducing exploration over noisy or irrelevant paths. As shown in Table~\ref{tab:ablation_study}, removing CLS enlarges the active set and leads to higher latency and tokens, indicating that CLS saves compute by avoiding low-value multilingual trajectories early.

{

\begin{figure}[t]
  \centering
  \includegraphics[width=\linewidth]{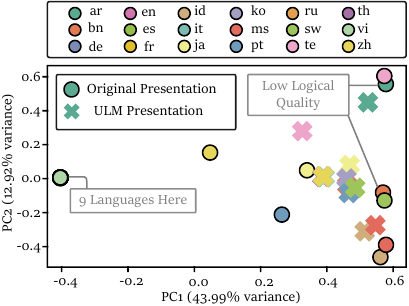}
  \caption{PCA projection of same-query embedding representations across 18 languages. Circles denote the original representations, while crosses indicate ULM-transformed representations in the unified logic space.}

  \vspace{-1em}
  \label{fig:logicunderstanding}
\end{figure}
}

\begin{figure*}[t]
  \centering
  \includegraphics[width=\linewidth]{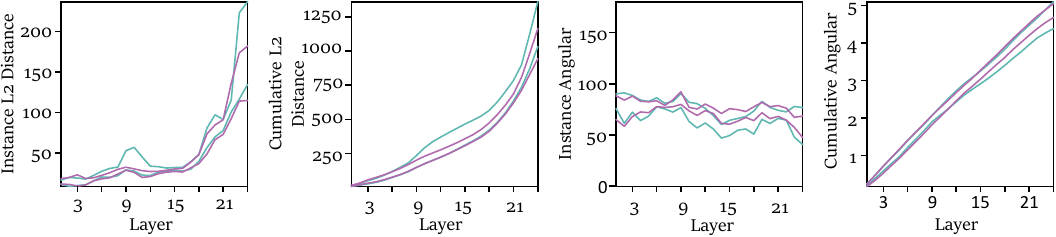}
  \caption{Layer-wise evolution of decoding embeddings measured by {L2 distance} and {Angular} across transformer layers.
  For each layer, we report the {sample-wise extrema} (min/max) to visualize the variation range.
  Purple denotes the model with ULM, while blue denotes the model without ULM.}
  \label{fig:decoding_analysis}
\end{figure*}

\paragraph{DCP saves compute via online pruning of low-quality trajectories.}
{DCP} further reduces computation through {early truncation}: it monitors trajectory quality online and stops paths that show clear signs of low utility. And they do not continue consuming the full decoding budget.
This strategy shrinks the active set early and focuses computation on the remaining competitive candidates during inference time. From Table~\ref{tab:ablation_study}, removing DCP substantially increases token usage and latency, while yielding only limited gains in final performance. These results suggest that many additional steps generated without DCP are largely redundant and rarely translate into better final votes.

\subsubsection{ULM can effectively unify cross-lingual logic representations.}
To assess how ULM unifies cross-lingual logical representations, we analyze (1) static alignment across languages and (2) dynamic decoding trajectories under ULM.

\noindent \textbf{ULM disentangles static language-specific variation.} To visualize static alignment, we extract logic-space embeddings for the same query across 18 languages at a fixed decoding step and project them into 2D using PCA. Figure~\ref{fig:logicunderstanding} shows that ULM of UL-XCoT disentangles surface-form variation, producing comparable, language-invariant representations across languages. After removing language-specific components, embeddings exhibit greater invariance: cross-lingual samples cluster tightly with consistent nearest neighbors. This indicates that the retained subspace encodes a shared logic state, not superficial differences.

\noindent \textbf{ULM can align dynamic reasoning trajectories.} Beyond static alignment, Figure~\ref{fig:decoding_analysis} tracks how hidden-state geometry evolves across layers when answering the same queries in different languages.

With ULM, trajectories show {a consistent distribution} across languages, reflecting shared evolution patterns in the logic space. Without ULM, trajectories diverge markedly, with greater sensitivity and larger geometric gaps due to linguistic difference. Thus, ULM removes superficial variation, enabling DCP's chain-of-thought pruning via {reliable geometric signals} in the unified logic space.

\begin{figure}[t]
  \centering
  \includegraphics[width=\linewidth]{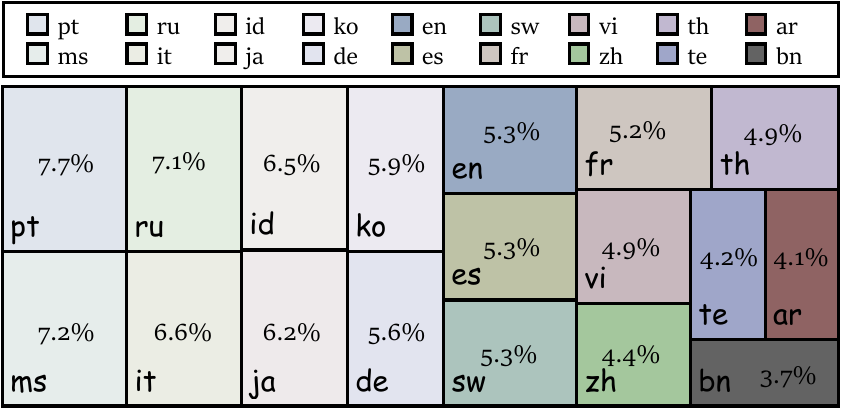}

  \caption{Distribution of languages selected by CLS, measured by the frequency of each language appearing in $\mathcal{L}_{\text{par}}(x)$ over the full evaluation suite.}
  \label{fig:clsselecting}
\end{figure}

\subsubsection{CLS can adaptively select appropriate languages without bias.}
As shown in Figure~\ref{fig:clsselecting}, we count how often each language appears in the CLS-selected set $\mathcal{L}_{\text{par}}(x)$ across the full evaluation suite. CLS does not collapse to a single language: each language contributes roughly $3.7\%\!\sim\!7.7\%$ of all selections (mean $\approx 5.6\%$), indicating broad coverage without a dominant bias. While a few languages (e.g., {pt/ms/ru/it/id}) are selected slightly more often, CLS still consistently includes lower-frequency languages (e.g., {bn/ar/te/zh}) for a non-trivial portion of inputs, reflecting query-adaptive selection rather than a fixed or heuristic language list.

\begin{figure*}[t]
  \centering
  {\captionsetup[subfigure]{labelformat=empty}}

  \begin{subfigure}[t]{0.33\textwidth}
    \centering
    \includegraphics[width=\linewidth]{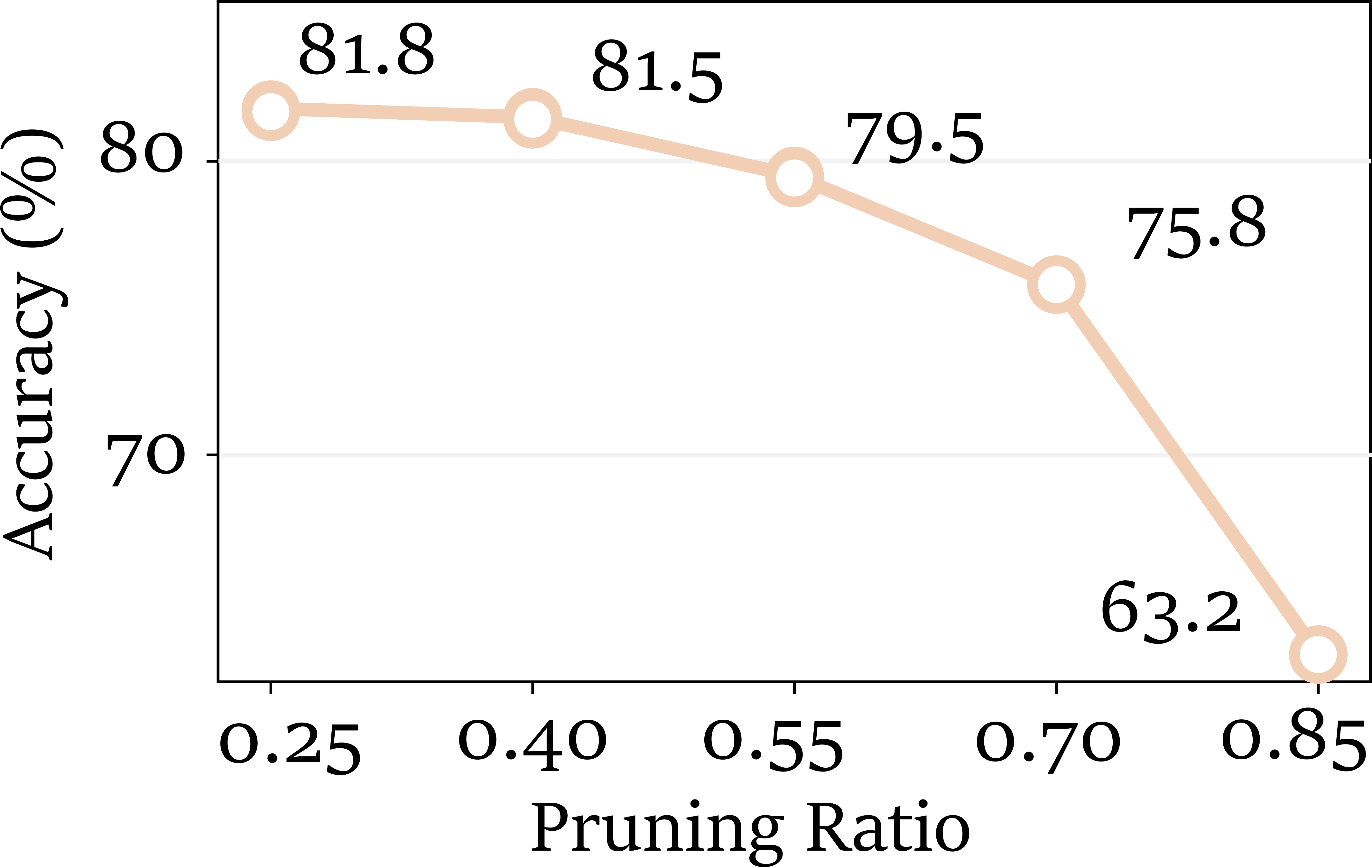}
  \end{subfigure}\hfill
  \begin{subfigure}[t]{0.318\textwidth}
    \centering
    \includegraphics[width=\linewidth]{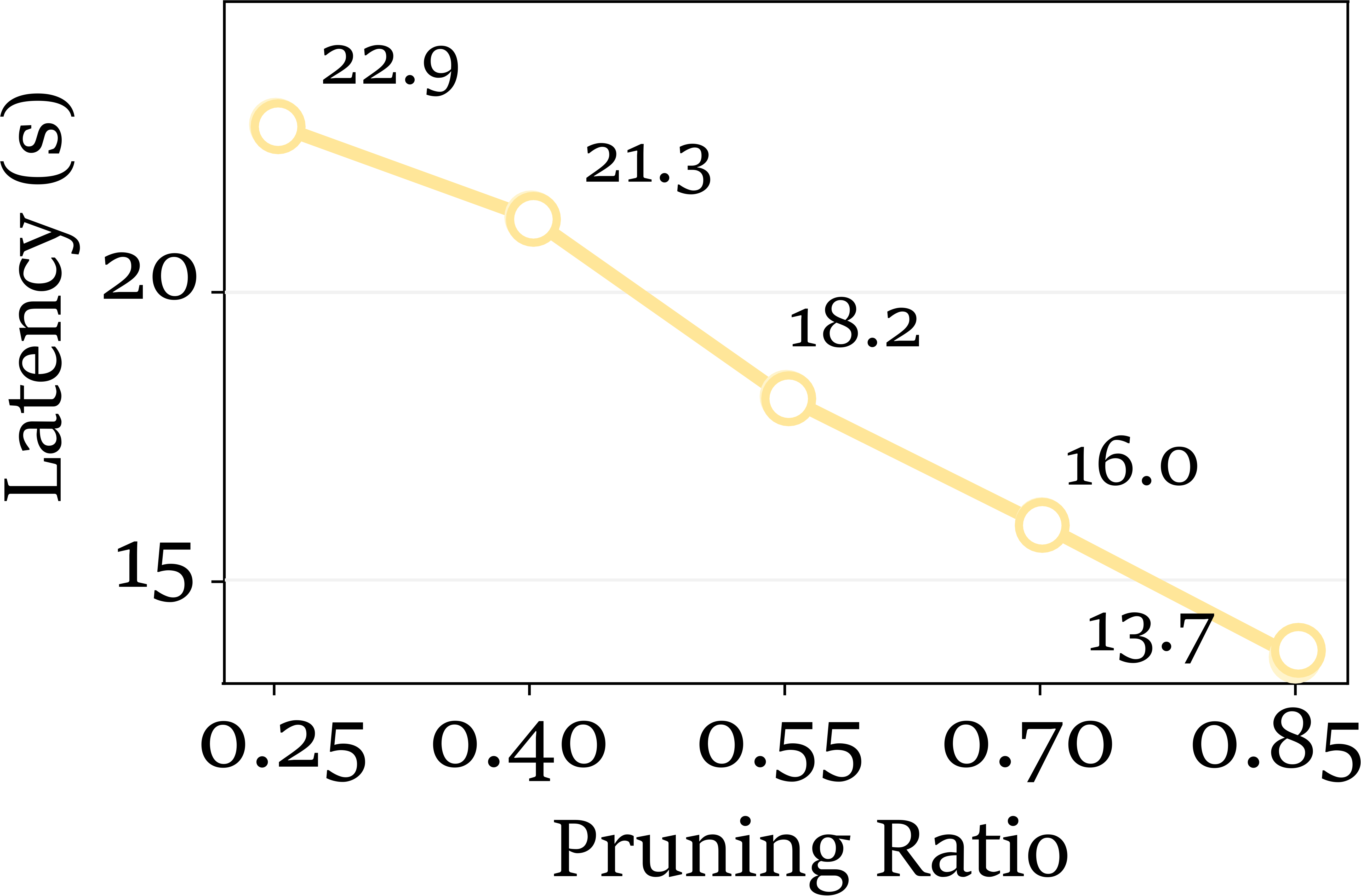}
  \end{subfigure}\hfill
  \begin{subfigure}[t]{0.35\textwidth}
    \centering
    \includegraphics[width=\linewidth]{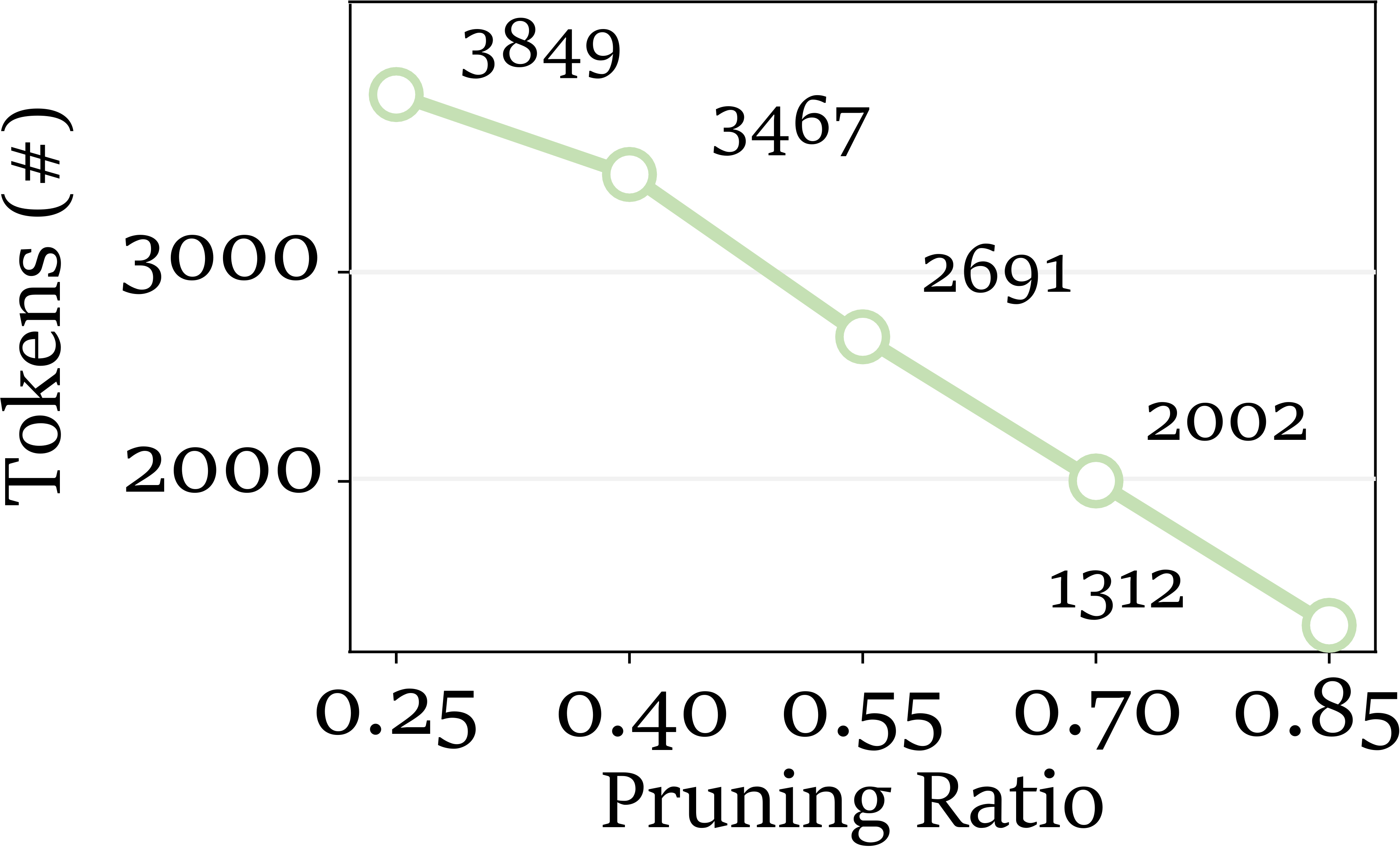}
  \end{subfigure}

  \caption{Impact of the pruning ratio $\rho$ on accuracy (left), latency (middle), and generated tokens (right).}
  \label{fig:pruning_analysis}
\end{figure*}

\subsubsection{DCP enables quality-aware pruning for efficiency.}

\paragraph{DCP can effectively balance performance and efficiency.}
To assess the impact of the pruning ratio $\rho$ in DCP, we vary $\rho$ from 0.0 to 0.9 and measure both accuracy and efficiency. As shown in Figure~\ref{fig:pruning_analysis}, for $\rho < 0.85$, higher $\rho$ slightly degrades accuracy but yields an almost linear reduction in latency and token usage, indicating that DCP removes low-confidence paths with minimal quality loss. Overall, a moderate $\rho$ of 0.55--0.70 provides the best trade-off, achieving substantial efficiency gains while largely preserving performance.

\begin{figure}[t]
  \centering
  \includegraphics[width=\linewidth]{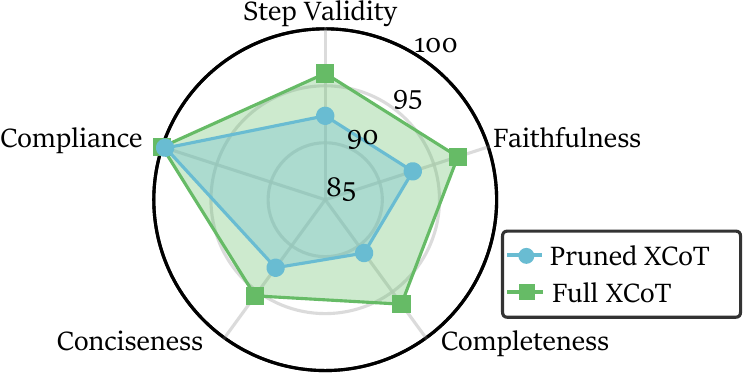}

  \caption{Quality comparison between trajectories pruned by DCP (Pruned XCoT) and those retained in Full XCoT, scored by an LLM judge over five criteria.}
  \label{fig:dcpxcot}
\end{figure}

\paragraph{DCP can truly prune low-quality paths.}
To verify DCP's pruning quality, we analyze a high-performing subset where both Pruned and Full XCoT achieve strong overall performance. As shown in Figure~\ref{fig:dcpxcot}, we use an LLM-as-a-judge\footnote{The prompt is provided in Appendix~\ref{app:prompt}} to score trajectories (0--100) on step validity, faithfulness, completeness, conciseness, and compliance. Pruned trajectories consistently score lower, with the largest drops in step validity and completeness, indicating weaker logical coherence and more missing intermediate reasoning.

%% file: section/related.tex
\section{Related Work}

\subsection{Cross-lingual Chain-of-Thought Reasoning}
Chain-of-Thought (CoT) prompting elicits explicit intermediate steps, improving reliability on multi-step problems~\citep{wei2022chain, kojima2022large, qin2024large,chen2025reasoningerasurveylong}. In cross-lingual settings, CoT can transfer across languages and strengthen in underrepresented languages as model scale grows~\citep{shi2022language, ghosh2025multilingual,barua2025longchainofthoughtreasoninglanguages,he2024scalinglawsmultilinguallanguage}. This motivates XCoT to go beyond prompt translation and exploit multilingual signals~\citep{huang2023not, wang2025large, chai2025xcot, ahuja2025efficientxlang}. Recently, XCoT studies instruction tuning for CoT transfer, distilling reasoning from high-resource to low-resource languages~\citep{Upadhayay2023, Kuulmets2024, chai2025xcot, Weihua2025}.

\subsection{Cross-lingual Chain-of-Thought Self-consistency}
Self-consistency improves CoT by sampling multiple paths and selecting the most consistent answer~\citep{wang2022self, aggarwal2023let, wang2025ranked}.

Cross-lingual self-consistent prompting extends this to XCoT paths~\citep{qin2023cross}, while AUTOCAP automates language selection and learns language-specific weights~\citep{zhang2024autocap}. Cross-lingual Tree-of-Thoughts performs multilingual search with branching reasoning and aggregation~\citep{ranaldi2024tree}. $L^2$ leverages multilingual unification learning and decoding-time interventions~\citep{chen2025less}. Best-of-L ranks multilingual candidates with a cross-lingual reward model for math reasoning~\citep{rajaee2025bestofl} and Multidimensional Consistency aggregates signals across input perturbations to improve robustness~\citep{lai2025multidimensional}.

Compared to these methods, our focus is efficiency-oriented allocation of test-time computation. We align intermediate states across languages in a unified logic space, which makes partial reasoning trajectories directly comparable during decoding and enables query-adaptive language selection together with online pruning.

%% file: section/conclusion.tex
\section{Conclusion}
We proposed UL-XCoT, an efficient cross-lingual reasoning framework that leverages a unified logic mechanism to better allocate compute resources. By selecting query-adaptive candidate languages and pruning inconsistent XCoT dynamically, {UL-XCoT} reduces redundancy at both the language and token levels while maintaining strong cross-lingual reasoning quality.

Experiments on PolyMath and MMLU-ProX-Lite show competitive accuracy with significantly higher efficiency and stable gains on a data-driven low-resource subset.

%% file: section/limitation.tex
\section*{Limitations}

Prior interpretability work suggests that transformer hidden states encode information-rich representations that can be meaningfully inspected and analyzed~\citep{Yang2024, ghandeharioun2024patchscopes, Skean2025}. Our method builds on this observation by leveraging hidden-state representations to compare cross-lingual understanding differences in a unified logic space and to monitor the quality of reasoning trajectories for online pruning. Therefore, UL-XCoT assumes white-box access to hidden states. Applicability to strict black-box LLM APIs remains to be validated.

%% file: section/acknowledgement.tex
\section*{Acknowledgement}
This work was supported by the National Natural Science Foundation of China (NSFC) via grants 92570120 and 62306342. This work was supported by the Scientific Research Fund of Hunan Provincial Education Department (24B0001). This work was sponsored by the Excellent Young Scientists Fund in Hunan Province (2024JJ4070), the Science and Technology Innovation Program of Hunan Province under Grant 2024RC3024. This study was also funded by the Open Project of the Text Computing and Cognitive Intelligence Ministry of Education Engineering Research Center (No. TCCI250101).

%% file: section/appendix.tex
\newpage
\appendix
\section{Mathematical Details of DCP}
\label{app:math}
\paragraph{Logic-space curvature signal.}
Building upon hidden-state based self-truncation ideas~\citep{wang2025sampling, zhang2025reasoningmodelsknowtheyre, bae2023fastrobustearlyexitingframework,chen2025llmssignaltheyreright}, we quantify the within-model stability of a trajectory by measuring the curvature of projected hidden states across layers at each decoding step. After warm-up, at decoding moment $t$, for each active language path $\ell$, let $x_{\ell}^{t}$ denote the current prefix. Let $m\in\mathcal{M}=\{m_s,\ldots,m_e\}$ be the monitored Transformer layers. We define the position-averaged projected hidden state: \[ h_{t,m}^{\ell} \coloneqq \hat{H}_{m}^{t}(x_{\ell}). \] \[ \hat{H}_{m}^{t}(x,\ell) = \frac{1}{|x_{\ell}^{t}|}\sum_{i=1}^{|x_{\ell}^{t}|}\tilde{H}^{t}_{i,m}(x,\ell). \] where $\tilde{H}^{t}_{i,m}(x,\ell)$ is the projected hidden state at token position $i$ and layer $m$.

We quantify layer-to-layer local changes in magnitude and direction. Define cosine similarity: \[ \cosim(u,v) \coloneqq \frac{u^\top v}{\lVert u \rVert\,\lVert v \rVert}. \] For each adjacent layer pair $(m-1,m)$, define the magnitude change: \[ \delta_M^{(t,m)}(\ell) \coloneqq \norm{h_{t,m}^{\ell}-h_{t,m-1}^{\ell}}. \] Define the angular change: \[ \delta_A^{(t,m)}(\ell) \coloneqq \arccos\!\big(\cosim(h_{t,m}^{\ell},h_{t,m-1}^{\ell})\big). \]

We normalize by the end-to-end (chord) change across layers at the same step $t$. Define: \[ \Delta_M^{t}(\ell) \coloneqq \norm{h_{t,m_e}^{\ell}-h_{t,m_s}^{\ell}}. \] \[ \Delta_A^{t}(\ell) \coloneqq \arccos\!\big(\cosim(h_{t,m_e}^{\ell},h_{t,m_s}^{\ell})\big). \] The layer-wise curvature ratios are: \[ r_M^{t}(\ell) \coloneqq \frac{\sum_{m=m_s+1}^{m_e}\delta_M^{(t,m)}(\ell)}{\Delta_M^{t}(\ell)}. \] \[ r_A^{t}(\ell) \coloneqq \frac{\sum_{m=m_s+1}^{m_e}\delta_A^{(t,m)}(\ell)}{\Delta_A^{t}(\ell)}. \]

Finally, we define a \textbf{Logic-space curvature signal} to measure the evolution of a path: \[ \kappa^{t}(\ell) \coloneqq r_M^{t}(\ell)-r_A^{t}(\ell). \]

\textbf{Divergence test.} To avoid false positives caused by a global drift shared by all paths, we implement an indicator $\mathbb{I}_{\kappa}t$ using both absolute and relative pairwise spread. Specifically, define the maximum absolute spread \[ \Delta_{\max}^{t} \coloneqq \max_{\ell\neq \ell'} \big|\kappa^{t}(\ell)-\kappa^{t}(\ell')\big|, \] the maximum relative spread \[ R_{\max}^{t} \coloneqq \max_{\ell\neq \ell'} \frac{\big|\kappa^{t}(\ell)-\kappa^{t}(\ell')\big|}
{\max\!\big(|\kappa^{t}(\ell)|,\ |\kappa^{t}(\ell')|,\ \delta\big)},
\] and the mean relative spread over all unordered pairs can be expressed as \[ R_{\mathrm{mean}}^{t} \coloneqq \mathbb{E}_{\ell\neq \ell'} \left[ \frac{\big|\kappa^{t}(\ell)-\kappa^{t}(\ell')\big|}
{\max\!\big(|\kappa^{t}(\ell)|,\ |\kappa^{t}(\ell')|,\ \delta\big)}
\right], \] where $\delta>0$ is a small constant for numerical stability. We declare step $t$ as \textbf{divergent} iff all three conditions hold: \[
\begin{aligned}
\mathbb{I}_{\kappa}^{t} &= \mathbb{I}\!\left[\Delta_{\max}^{t}>\varepsilon_{\mathrm{abs}}\right]\cdot \mathbb{I}\!\left[R_{\max}^{t}>\varepsilon_{\mathrm{rel}}\right] \\ &\quad\cdot \mathbb{I}\!\left[R_{\max}^{t}\ge \gamma\,R_{\mathrm{mean}}^{t}\right].
\end{aligned}
\]

\paragraph{Divergence detection and scoring.}
Let $S_t^*$ denote the set of active paths at decoding step $t$ and $N_t \!=\! |S_t|$. For each path $\ell \in S_t$, let $\kappa^{t}(\ell)$ be its divergence descriptor at step $t$. We start monitoring after a warm-up period and define the first post-warm-up divergence step as \[ c \coloneqq \min \{\, t \ge T_{\mathrm{warm}} \mid \mathbb{I}_{\kappa}^{t} = 1 \,\}. \] We only score paths within a fixed window \[ t \in [c,\, c+\tau]. \]

\paragraph{Per-step point assignment.}
We set $K_t'\coloneqq \min(K',N_t)$ and assign each path $\ell\in S_t$ a binary point $\texttt{score}(S_t|x_\ell,\ell')\in\{0,1\}$:
\begin{equation}
  \label{eq:prune_score}
  \begin{aligned}
  &\begin{cases}
  \mathbb{I}[\ell\in R_t], & \mathbb{I}_{\kappa}^{t}=0 \;,\\
  \mathbb{I}[\ell\in W_t], &  \mathbb{I}_{\kappa}^{t}=1 \;,
  \end{cases}
  \end{aligned}
\end{equation}

where, if step $t$ is non-divergent, we sample $R_t\subseteq S_t$ uniformly at random without replacement with $|R_t|=K'$. If step $t$ is divergent, we keep the $K'$ most central paths by minimizing the average distance to the cohort in the divergence space: \[ g^{t}(\ell)\coloneqq \frac{1}{\max(1,N_t-1)}\sum_{\ell'\in S_t\setminus\{\ell\}} \big|\kappa^{t}(\ell)-\kappa^{t}(\ell')\big| \] \[ W_t \coloneqq \operatorname{TopK'}_{\,\ell\in S_t}\!\big(-g^{t}(\ell)\big). \] Equivalently, $W_t$ contains the $K_t'$ paths with the smallest $g^{t}(\ell)$, with ties broken arbitrarily.

\begin{figure*}[t]
  \centering
  \begin{subfigure}[t]{\textwidth}
    \centering
    \includegraphics[width=\linewidth]{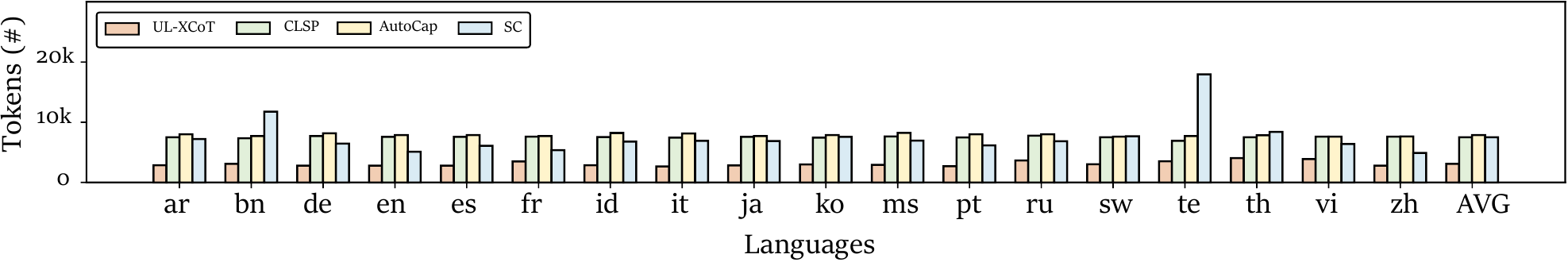}
    \caption{PolyMath-low}
    \label{fig:tokens_low}
  \end{subfigure}\hfill
  \begin{subfigure}[t]{\textwidth}
    \centering
    \includegraphics[width=\linewidth]{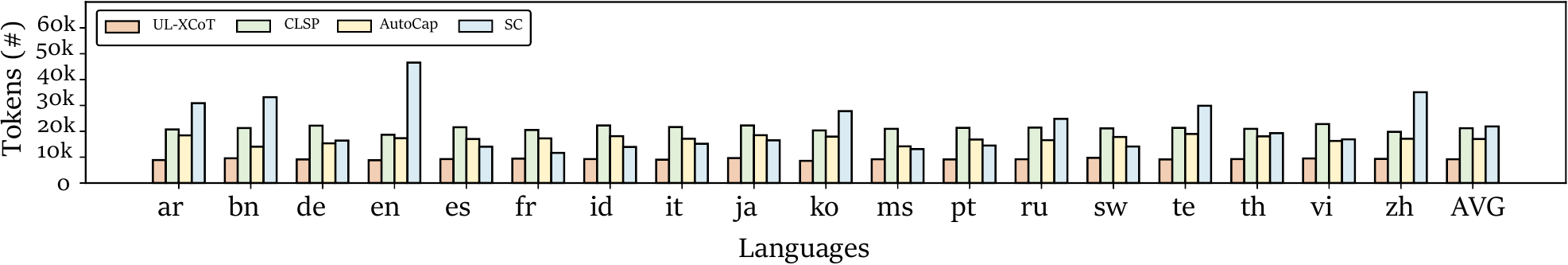}
    \caption{PolyMath-medium}
    \label{fig:tokens_medium}
  \end{subfigure}

  \vspace{2mm}

  \begin{subfigure}[t]{\textwidth}
    \centering
    \includegraphics[width=\linewidth]{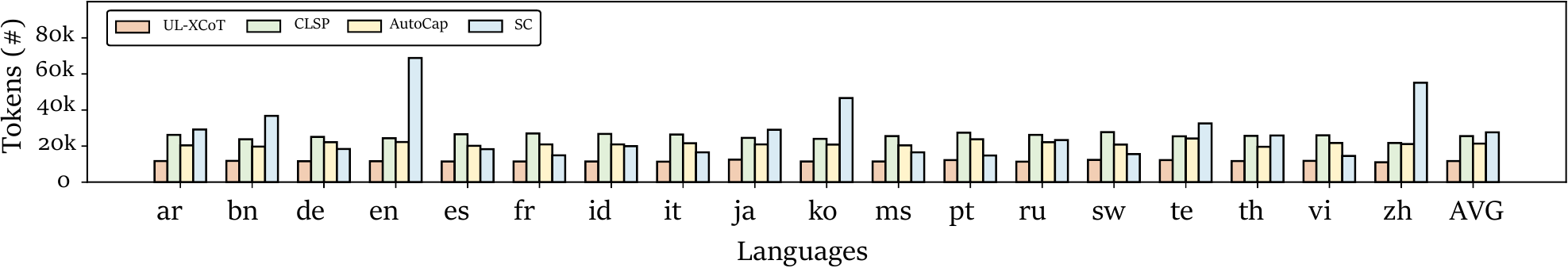}
    \caption{PolyMath-high}
    \label{fig:tokens_high}
  \end{subfigure}\hfill
  \begin{subfigure}[t]{\textwidth}
    \centering
    \includegraphics[width=\linewidth]{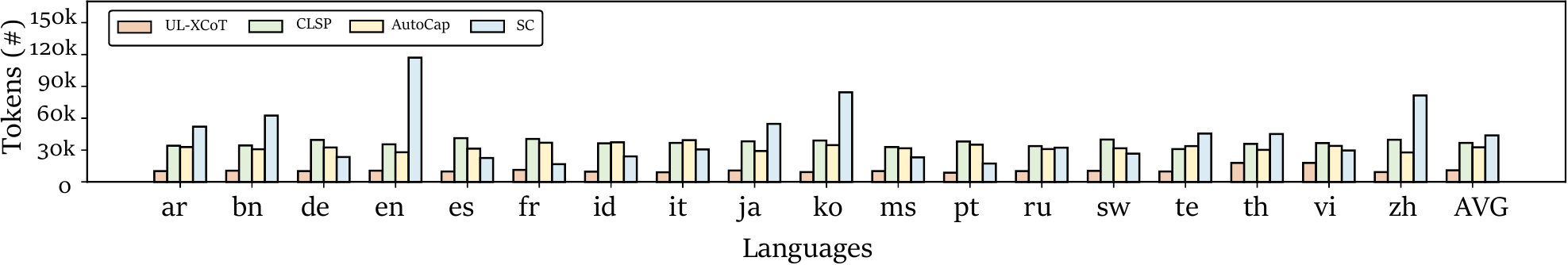}
    \caption{PolyMath-top}
    \label{fig:tokens_top}
  \end{subfigure}

  \caption{Decoding token cost during generation across PolyMath difficulty levels.}
  \label{fig:tokens_all}
\end{figure*}

\begin{figure*}[t]
  \centering

  \begin{subfigure}[t]{\textwidth}
    \centering
    \includegraphics[width=\linewidth]{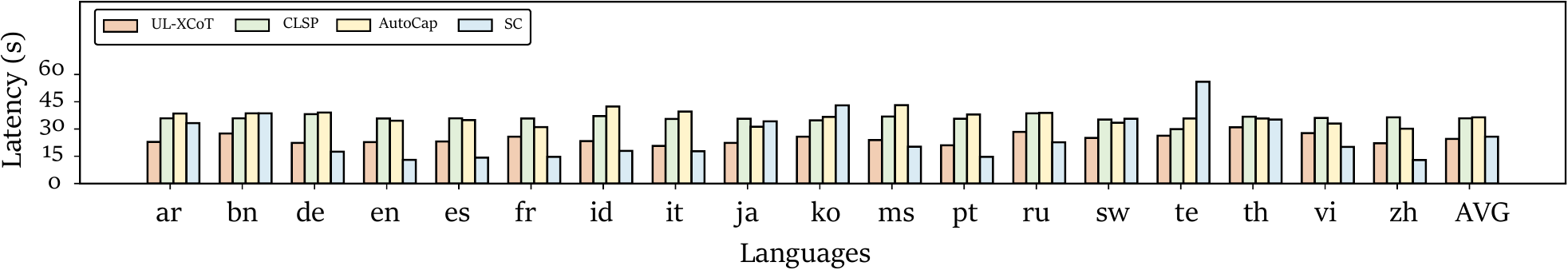}
    \caption{PolyMath-low}
    \label{fig:latency_low}
  \end{subfigure}\hfill
  \begin{subfigure}[t]{\textwidth}
    \centering
    \includegraphics[width=\linewidth]{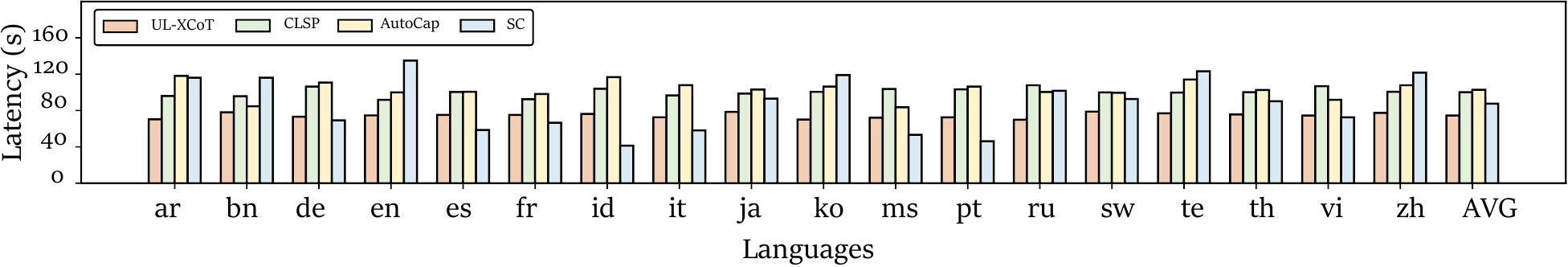}
    \caption{PolyMath-medium}
    \label{fig:latency_medium}
  \end{subfigure}

  \vspace{2mm}

  \begin{subfigure}[t]{\textwidth}
    \centering
    \includegraphics[width=\linewidth]{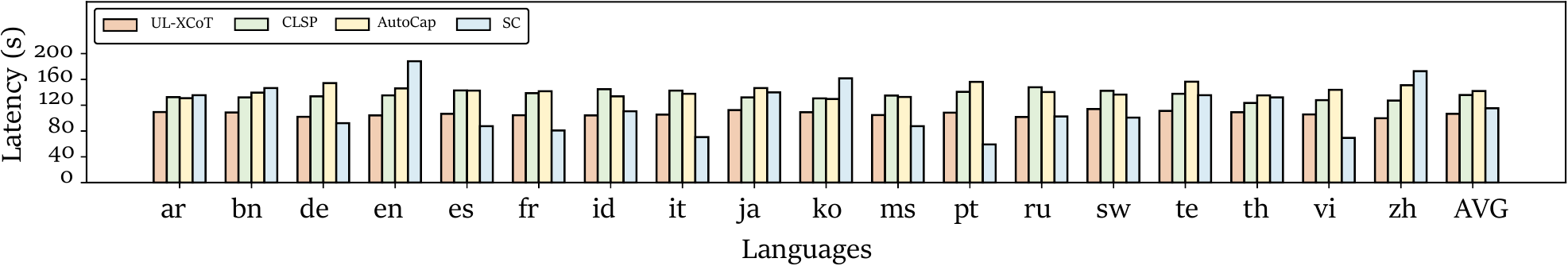}
    \caption{PolyMath-high}
    \label{fig:latency_high}
  \end{subfigure}\hfill
  \begin{subfigure}[t]{\textwidth}
    \centering
    \includegraphics[width=\linewidth]{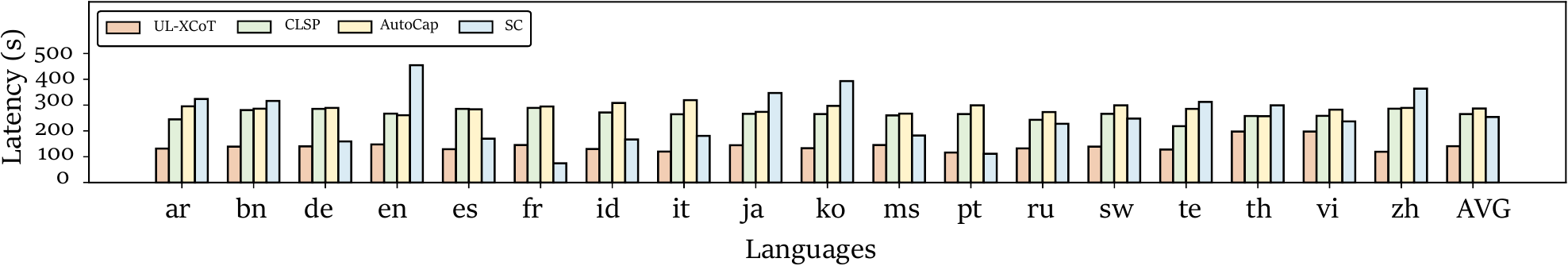}
    \caption{PolyMath-top}
    \label{fig:latency_top}
  \end{subfigure}

  \caption{Decoding wall-clock latency during generation across PolyMath difficulty levels.}
  \label{fig:latency_all}
\end{figure*}

\section{Detailed Efficiency Analysis}
This appendix analyzes the efficiency results in Sec.~\ref{sec:superior_efficiency} by PolyMath difficulty level, providing additional evidence for the {less languages, less tokens} motivation.
\label{app:efficiency}

\subsection{Token cost across difficulty levels.}
Figure~\ref{fig:tokens_all} reports the decoding token cost over four PolyMath difficulty levels, broken down by language. As difficulty increases from low to top, all methods exhibit higher token consumption, indicating longer reasoning traces on harder problems. Across all languages and levels, UL-XCoT consistently incurs fewer tokens than the baselines, showing that query-adaptive language selection and online trajectory pruning can reduce unnecessary generation without relying on extra sampling. The gap becomes more evident on harder subsets (high/top), where baseline decoding tends to produce longer and more variable traces, while UL-XCoT keeps token usage more stable and cost-efficient.

\subsection{Wall-clock latency across difficulty levels.}
Figure~\ref{fig:latency_all} reports decoding wall-clock latency during generation across four PolyMath difficulty levels, broken down by language. Latency increases monotonically from low to top for all methods, reflecting the longer and more compute-intensive reasoning required by harder problems. Across nearly all languages and all levels, UL-XCoT consistently achieves the lowest latency and the best AVG, indicating a more efficient decoding profile than CLSP, AUTOCAP, and SC. Notably, the advantage becomes more pronounced on high/top, where baseline methods exhibit larger variance and occasional long-tail spikes, while UL-XCoT remains more stable. This improvement is attributed to selecting candidate languages before decoding and dynamically pruning low-quality XCoT paths early, thereby avoiding unnecessary generation early and reducing end-to-end serving time.

\begin{table*}[!h]
  \centering
  \caption{Difficulty considered performance on PolyMath across 18 languages and four difficulty levels.}
  \label{tab:dw-acc_polymath}
  \resizebox{\textwidth}{!}{
  \begin{tabular}{l *{24}{c}}
  \toprule

  \textbf{DW-ACC} & \textbf{ar} & \textbf{bn} & \textbf{de} & \textbf{en} & \textbf{es} & \textbf{fr} & \textbf{id} & \textbf{it} & \textbf{ja} & \textbf{ko} & \textbf{ms} & \textbf{pt} & \textbf{ru} & \textbf{sw} & \textbf{te} & \textbf{th} & \textbf{vi} & \textbf{zh} & \textbf{AVG} \\
  \midrule
  \texttt{CoT}~\citep{wei2022chain}
  & 10.4 & 7.1 & 12.8 & 11.8 & 14.1 & 13.0 & 10.9 & 11.0 & 11.3 & 9.3 & 9.3 & 13.3 & 12.3 & 2.1 & 2.8 & 10.6 & 11.4 & 14.0 & 10.4 \\
  \texttt{CLP}~\citep{qin2023cross}
  & 9.0 & 7.9 & 11.9 & 13.4 & 10.3 & 13.3 & 9.9 & 11.6 & 13.8 & 10.7 & 12.5 & 11.1 & 12.1 & 11.0 & 9.2 & 8.1 & 11.1 & 11.4 & 11.0 \\
  \midrule
  \texttt{SC}~\citep{wang2022self}
  & \textbf{18.9} & 12.5 & 14.6 & \textbf{20.2} & 16.5 & 16.7 & 14.7 & 17.6 & 11.4 & 19.4 & 14.4 & 15.5 & 17.4 &  4.7 &  6.2 & 14.5 & 15.5 & \textbf{22.6} & 15.2 \\
  \texttt{CLSP}~\citep{qin2023cross}
  & 18.2 & 16.9 & \textbf{20.8} & 19.4 & 16.8 & 18.8 & \textbf{20.7} & 16.7 & \textbf{19.7} & \textbf{20.5} & 18.5 & 16.7 & 17.7 & 19.2 & \textbf{18.8} & 18.5 & 18.6 & 17.7 & 18.6 \\
  \texttt{AUTOCAP}~\citep{zhang2024autocap}
  & 18.3 & 18.3 & 17.8 & 17.3 & 18.3 & \textbf{19.4} & 18.8 & 16.1 & 17.9 & 20.0 & \textbf{20.5} & \textbf{18.3} & 16.3 & 17.1 & 17.9 & 15.0 & 17.7 & 18.2 & 18.2 \\
  \texttt{ST-BoN}~\citep{wang2025sampling}
  & 13.8 & 12.4 & 11.7 & 12.7 & 12.2 & 14.2 & 13.4 & 14.2 & 13.9 & 13.4 & 14.7 & 14.1 & 12.6 & 12.0 & 12.3 & 15.0 & 14.3 & 13.1 & 13.3 \\
  \rowcolor[rgb]{0.929, 0.961, 0.980}
  \texttt{\textbf{UL-XCoT}}
  & 18.5 & \textbf{19.1} & 19.3 & 19.7 & \textbf{19.1} & 19.1 & 20.0 & \textbf{18.1} & 17.8 & 17.3 & 19.3 & 17.4 & \textbf{20.1} & \textbf{21.8} & 18.0 & \textbf{20.0} & \textbf{19.7} & 18.0 & \textbf{19.0} \\
  \bottomrule
  \end{tabular}}
\end{table*}
\newpage
\section{Overall Performance on PolyMath}
\label{app:polymath}
Table~\ref{tab:dw-acc_polymath} provides the \textbf{PolyMath-Full} summary in terms of \textbf{DW-ACC}, which is the primary effectiveness metric used in our evaluation protocol. Different from Table~\ref{tab:acc_polymath_the_whole} that reports accuracy separately for each difficulty level, DW-ACC \textbf{aggregates} performance over the four levels into a single difficulty-aware score, enabling a compact comparison of overall effectiveness across languages.

Overall, UL-XCoT achieves the best {AVG} DW-ACC, and remains consistently strong across the 18-language suite. This result directly supports that our efficiency-oriented framework substantially reduces inference cost, it stays competitive in difficulty-weighted accuracy on PolyMath-Full. In particular, the gains are not confined to a single high-resource language, but are distributed across languages, suggesting that the proposed CLS and DCP strategy guarantees cross-lingual reasoning quality under the same decoding budget.

\clearpage
\newpage

\section{Prompt Templates}
\label{app:prompt}
\subsection{Concise Reasoning Prompt}
We faithfully implement the reasoning setting using the following prompt, which enforces a concise thinking process and enables a clearer analysis of performance under such a regime~\citep{renze2024benefits,xu2025chain}.

Specifically, it (i) fixes the reasoning language to \texttt{\textless LANG\_NAME\textgreater} to avoid cross-lingual leakage, (ii) caps the reasoning to at most \texttt{\textless STEP\_NUM\textgreater} numbered steps to control verbosity and token budget, and (iii) restricts the final output to a single boxed answer outside the \texttt{\textless think\textgreater} block, ensuring a clean separation between intermediate reasoning and the model's final response. This standardized format makes results comparable and isolates efficiency gains attributable to the decoding strategy rather than prompt-induced length differences.

\begin{figure*}[t]
  \centering
  \begin{tcolorbox}[
    colback=green!3!white,
    colframe=green!45!black,
    width=\textwidth,
    arc=4pt,
    boxrule=0.5pt,
    title=\textbf{concise-reasoning template}
  ]
  
      \setlength{\parindent}{0pt}
      \setlength{\parskip}{0.8ex}

      You are an expert in mathematical / geometric reasoning.

      Think strictly in <LANG\_NAME> step by step. Do not use any other language.

      \textbf{Format:}

      \texttt{<think>}

      Step 1: ...

      Step 2: ...

      ...

      Step N: ...

      \texttt{</think>}

      \texttt{\detokenize{$\boxed{FINAL\_ANSWER}$}}

      \textbf{Hard Rules:}

      1) All intermediate reasoning MUST be inside a single \texttt{<think>...</think>} block, written only in <LANG\_NAME>.

      2) At most <STEP\_NUM> numbered steps. Be concise and avoid repetition.

      3) Outside \texttt{</think>} you may output ONE line only: \texttt{\detokenize{$\boxed{FINAL\_ANSWER}$}}.

      4) Do NOT restate the problem. Do NOT add any explanation, comments, or extra text after the boxed answer.

      5) If the result is an expression, keep it simplified. If numeric, give an exact value when possible.

      \textbf{Question:}

      <QUERY>

      \textbf{Nnotes:}

      - Use standard math notation. Keep symbols/variables as-is.

      - If you reach a conclusion early, stop immediately and output the boxed answer.

  \end{tcolorbox}\vspace{-16pt}
\end{figure*}

\subsection{Quality Judge Prompt}

We use an \textbf{LLM-as-a-judge} prompt to score each candidate trajectory in a
{structured} and {machine-readable} manner~\citep{zheng2023judging,li2024llms}.
We use GPT\textendash4o mini as the LLM in our experiments~\citep{openai_gpt4omini_blog}. The judge takes the \texttt{question}, \texttt{reference\_answer}, \texttt{candidate\_answer}, and the candidate CoT as input, and is constrained to output \textbf{only} a JSON object matching a fixed schema. It assigns integer scores in $[0,100]$ on six dimensions: correctness (exact match to the reference answer), step validity (logical soundness without jumps), faithfulness (no hallucinated facts), completeness (covers key constraints), conciseness (non-redundant), and compliance (language constraints). The final \texttt{overall} score is a weighted average with the largest weight on correctness.
\begin{figure*}[t]
  \centering
  \begin{tcolorbox}[
    colback=green!3!white,
    colframe=green!45!black,
    width=\textwidth,
    arc=4pt,
    boxrule=0.5pt,
    title=\textbf{quality-judge-prompt}
  ]
  
      \setlength{\parindent}{0pt}
      \setlength{\parskip}{0.8ex}

      You are a strict grader. Output ONLY JSON matching the schema.

      Score scale (IMPORTANT):

      - All scores are INTEGERS from 0 to 100 (100 is best).

      - 0–20: very poor, 40–60: mediocre, 70–80: good, 85–95: very good, 96–100: near-perfect.

      - DO NOT use a 1–5 scale.
      
      \textbf{Question:}

      <question>
      
      \textbf{Reference Answer:}

      <reference\_answer>
      
      \textbf{Candidate Answer:}

      <candidate\_answer>
      
      \textbf{Candidate Reasoning / CoT:}

      <candidate\_cot>
      
      \textbf{Dimension rules:}

      - correctness: 100 if candidate answer matches reference answer (allow trivial formatting), else 0.

      - step\_validity: penalize jumps/invalid inference; 100 means each step is logically justified.

      - faithfulness: penalize invented facts not in question or derivable; 100 means fully grounded.

      - completeness: 100 means all key constraints/calculations covered.

      - conciseness: 100 means no redundancy; lower if repetitive.

      - compliance: 100 means follows required format/language constraints.

      overall: weighted average (correctness has the largest weight).

  \end{tcolorbox}\vspace{-16pt}
\end{figure*}